\documentclass{article}
\usepackage{arxiv}
\renewcommand{\headeright}{}
\renewcommand{\undertitle}{}

\usepackage[utf8]{inputenc} 
\usepackage[T1]{fontenc}    
\usepackage{hyperref}       
\usepackage{url}            
\usepackage{booktabs}       
\usepackage{amsfonts}       
\usepackage{nicefrac}       
\usepackage{microtype}      
\usepackage{float}          
\usepackage{tikz}

\usepackage{lipsum}         
\usepackage{booktabs}
\usepackage{longtable}
\usepackage{multirow}
\usepackage{array}
\usepackage{graphicx}
\usepackage[round]{natbib}
\usepackage{doi}
\usepackage{booktabs}   
\usepackage{tabularx}   
\usepackage[utf8]{inputenc}
\usepackage{amsmath}    
\usepackage{geometry}
\usepackage{longtable}
\usepackage{multirow}
\usepackage{array}
\usepackage{xcolor}
\usepackage{booktabs}
\usepackage{longtable}
\usepackage{pdflscape}
\usepackage{array}
\usepackage{tabularx}
\usepackage{amssymb}
\usepackage{booktabs}
\usepackage{multirow}
\usepackage{adjustbox}

\newcolumntype{C}[1]{>{\centering\arraybackslash}p{#1}}
\usepackage{rotating}  
\usepackage{pdflscape} 
\definecolor{graybg}{HTML}{F5F5F5}
\usepackage[table]{xcolor}
\usepackage{colortbl}
\definecolor{dimgray}{HTML}{F0F2F5}
\usepackage{graphicx}
\usepackage{subcaption}   
\usepackage{authblk}  

\usepackage{caption}
\usepackage{ltablex} 
\keepXColumns 

\title{MedBench v5: A Dynamic, Process-Oriented, and Hallucination-Aware Benchmark for Clinical Multimodal Models}

\author[1]{Jinru Ding%
  \thanks{Co-first author. dingjinru@pjlab.org.cn}%
}
\author[1]{Chuchu Jiang%
  \thanks{Co-first author. jiangchuchu@pjlab.org.cn}%
}
\author[1]{Lu Lu%
  \thanks{Co-first author. lulu@pjlab.org.cn}%
}
\author[1]{Wenrao Pang%
}
\author[1]{Mouxiao Bian%
}
\author[1]{Zhuangzhi Gao%
}
\author[1]{Jiangyuan Chen%
}
\author[1]{Xinwei Peng %
}

\author[1]{Ruiyao Chen%
}
\author[1]{Sijie Ren%
}
\author[1]{Renjie Lu%
}
\author[1]{Yun Zhong%
}
\author[1]{Bin Han%
}
\author[1]{Meiling Liu%
}
\author[1]{Jie Xu%
  \thanks{Corresponding author. xujie@pjlab.org.cn}%
}

\affil[1]{Shanghai Artificial Intelligence Laboratory}

\date{}
\renewcommand{\shorttitle}{}

\hypersetup{
pdftitle={A template for the arxiv style},
pdfsubject={q-bio.NC, q-bio.QM},
pdfauthor={MedBench Group},
pdfkeywords={Mutimodel Model Evaluation, Information-Flow Stressor, Audit Protocol, Hallucination Propagation Monitoring},
}

\begin{document}
\maketitle

\begin{abstract}
Existing medical AI benchmarks lack process visibility, atomic skill evaluation, and integrated hallucination detection. We introduce MedBench v5, a redesigned benchmark for clinical multimodal models (language, vision–language, and agent systems) that moves from static QA to dynamic, process-oriented evaluation. MedBench v5 features: (1) a dual-dimensional framework combining Clinical Cognitive Responsiveness (13 sub-dimensions) and Medical Atomic Skills (4 agent environments), covering 63 tasks; (2) three switchable information-flow stressors (omission, contradiction, evidence delay) for factorized degradation analysis; (3) a dynamic process audit protocol with five reasoning nodes that produces model-specific failure fingerprints; (4) hallucination propagation monitoring across initiation, propagation, anchoring, and contradiction interaction—capturing silent hallucination. Experiments on frontier models show that strong overall task performance does not guarantee process stability: stressors mainly disrupt contradiction detection, diagnosis updating, hallucination propagation, and contradiction-based self-correction, while final evidence grounding can remain superficially stable. MedBench v5 provides a unified infrastructure for capability profiling, controllable stress testing, process auditing, and hallucination trajectory analysis in clinical AI evaluation.
\end{abstract}


\section{Introduction}
Large language models (LLMs) and multimodal foundation models have shown growing potential in medical applications, including online pre-consultation, clinical documentation, intelligent follow-up, patient education, medical question answering, and clinical decision support \citep{singhal2023large, jung2025large, liu2024using, song2025large, aydin2024large, acosta2022multimodal, moor2023foundation}. However, real-world medical practice is not a static question-answering task. It is inherently dynamic and iterative: physicians must reason under uncertainty, actively elicit missing history, reconcile conflicting evidence, update diagnostic hypotheses, and make sequential decisions as new information becomes available \citep{sooknanan2019not, ball2015improving, meyer2021patient, weinstein2017diagnosing, thampy2019assessing, mccoy2025assessment}.

In contrast, most medical LLM benchmarks have predominantly adopted a static, single-turn question-answering (QA) paradigm, where the model receives a complete case description or exam-style prompt and produces a single final answer. Representative examples include MedQA \citep{zhang2021medq}, CMExam \citep{liu2023benchmarking}, MedMCQA\citep{pal2022medmcqa}, and CBLUE \citep{zhang2022cblue}. Although these benchmarks have played an important role in measuring medical knowledge, language understanding, and exam-style reasoning, they provide limited evidence of whether a model can operate safely in realistic clinical workflows \citep{kim2025questioning, sun2025beyond, bielick2026moving}. Recent studies have increasingly highlighted this mismatch between static benchmark performance and clinical readiness \citep{jiang2026beyond, chen2025beyond, wu2026bridge}. For example, systematic evidence suggests a persistent knowledge--practice gap: models that achieve strong performance on knowledge-based medical exams may perform substantially worse on practice-oriented or safety-critical tasks \citep{gong2025knowledge}. Similarly, when evaluation shifts from static cases to multi-turn or agent-based clinical interaction, diagnostic performance can degrade markedly, revealing failures that are hidden by single-turn QA evaluation \citep{sangwon2025evaluating, schmidgall2026agentclinic}.

Recognizing these limitations, recent benchmarks have begun to move toward practice-oriented and interactive evaluation \citep{liu2025interactive}. One line of work introduces multi-turn diagnostic dialogue tasks in which models must actively ask questions, gather missing information, and decide when sufficient evidence has been collected. Benchmarks such as MediQ \citep{li2024mediq}, Q4Dx \citep{werthaim2026benchmark}, and VivaBench \citep{chiu2026simulating} evaluate whether models can conduct sequential information seeking rather than merely answer a fully specified prompt. Another line of work studies robustness under incomplete, hidden, or adversarial patient information. For instance, MedConceal \citep{han2026medconceal} evaluates hidden-concern reasoning in medical dialogue, while MedDialBench \citep{luo2026meddialbench} examines diagnostic robustness under parametrically controlled non-cooperative patient behaviors. In parallel, agentic simulation environments\citep{liu2026medchain} such as MedAgentBench \citep{jiang2025medagentbench}, AgentClinic \citep{schmidgall2026agentclinic}, ClinEnv \citep{lu2026clinenvinteractivemultistagelong}, and MeDxAgent \citep{sanghvi2026medxagentmultiagentconsultationinteractive} embed models in more realistic clinical workflows \citep{yan2026clinicallab}, requiring them to retrieve information, interact with simulated patients or electronic health records, consult tools or specialist agents, and perform sequential clinical actions.

Safety evaluation has also become increasingly important as medical LLMs move closer to deployment \citep{asgari2025framework, roustan2025clinicians}. In particular, hallucination is a critical risk in clinical settings, because unsupported or fabricated claims may appear plausible while leading to unsafe diagnostic or therapeutic decisions \citep{zhu2025can, kim2025medical}. Existing hallucination-oriented benchmarks, including Med-HALT \citep{pal2023med}, MedHallu \citep{pandit2025medhallu}, and multimodal benchmarks such as Med-HallMark \citep{chen2024detecting} and MedVH \citep{gu2026medvh}, evaluate whether models can detect or avoid factual inaccuracies in medical responses. These efforts provide valuable tools for measuring factual reliability, especially at the response or final-answer level.

Despite this progress, existing practice-oriented benchmarks remain largely observational rather than diagnostic \citep{chen2025beyond}. They can reveal that model performance degrades in interactive or safety-critical settings, but they often cannot explain where and why the degradation occurs \citep{sun2025beyond, zhou2025automating, wang2025scores}. We identify four key limitations. First, many benchmarks still rely on end-to-end outcome scores, making it difficult to localize failures to specific reasoning stages such as recognizing missing information, asking appropriate follow-up questions, detecting contradictions, updating diagnoses, or grounding conclusions in evidence \citep{qiu2025quantifying}. Second, although recent interactive benchmarks introduce incomplete or adversarial information, few provide a controllable information-flow design that systematically toggles omission, contradiction, and delay to distinguish general task difficulty from specific cognitive failure modes \citep{li2026beyond, yan2025llm}. Third, existing evaluations often under-specify the atomic operational skills required by clinical AI systems, such as structured data manipulation \citep{shi2024ehragent}, retrieval-augmented reasoning \citep{xiong2024benchmarking}, long-horizon research synthesis \citep{huang2025deep}, and adversarial safety defense \citep{zhang2025agent} in executable or sandboxed environments. Fourth, hallucination evaluation is usually treated as a standalone factuality task, decoupled from the main clinical reasoning trajectory \citep{asgari2025framework}. As a result, current benchmarks rarely trace how unsupported facts emerge, persist across turns, interact with contradictions, and eventually contaminate final diagnostic conclusions \citep{lu2026mhb, yang2026clinhallu}.

To address these gaps, we introduce \textbf{MedBench v5}, a holistic benchmark for clinical multimodal model evaluation that moves beyond static QA toward dynamic, process-oriented, and hallucination-aware assessment. MedBench v5 combines a dual-dimensional capability framework with a stress-audit-tracing protocol, enabling both broad capability coverage and fine-grained failure attribution. Specifically, MedBench v5 introduces four key components:

\begin{itemize}
    \item \textbf{Dual-Dimensional Evaluation Framework.}
    We organize clinical model evaluation along two complementary dimensions: \textit{Clinical Cognitive Responsiveness} (CCR) and \textit{Medical Atomic Skills} (MAS). CCR covers 13 clinical capability dimensions spanning medical QA, natural language understanding and generation, clinical reasoning, multimodal perception, decision support, interaction, memory, tool use, safety, and multi-agent collaboration. MAS further instantiates four executable agent-based environments---DataAgent, RAGAgent, DeepResearch, and SafetyAgent---to evaluate structured data interaction, retrieval-augmented generation, long-horizon evidence synthesis, and adversarial safety defense. Together, these dimensions define 17 capability areas across 63 clinical tasks.

    \item \textbf{Switchable Information-Flow Stressors.}
    We design three independently togglable stressors---\textit{information omission}, \textit{contradiction injection}, and \textit{evidence delay}---to systematically perturb clinical information flow. By comparing no-stress, single-stressor, and multi-stressor conditions, this design enables controlled attribution of performance degradation to specific information-flow disruptions rather than treating interactive difficulty as an undifferentiated source of error.

    \item \textbf{Dynamic Five-Node Process Audit.}
    We establish a five-node audit protocol that evaluates model behavior across \textit{information gap detection}, \textit{follow-up strategy}, \textit{contradiction detection}, \textit{diagnosis update}, and \textit{evidence grounding}. Instead of scoring only the final answer, the audit records process-level behavioral traces and generates a reasoning failure profile for each model, revealing where the clinical reasoning chain breaks under different stress conditions.

    \item \textbf{Hallucination Propagation Monitoring.}
    Complementing the five-node audit, we monitor hallucination propagation throughout the multi-turn trajectory. This module tracks four progressive dimensions---initiation, propagation, anchoring, and hallucination--contradiction interaction---to quantify when unsupported claims first appear, whether they persist or cross-contaminate later reasoning, whether they become anchored in the final diagnostic evidence chain, and whether explicit contradictions suppress or induce further fabrication.
\end{itemize}

By integrating broad capability evaluation, executable atomic skill testing, controllable information-flow stressors, process-level auditing, and trajectory-level hallucination monitoring, MedBench v5 provides a clinically grounded and diagnostically transparent benchmark for medical LLMs and multimodal clinical AI systems. Rather than merely asking whether a model produces the correct final answer, MedBench v5 evaluates how the answer is reached, where the reasoning process fails, and how unsupported information propagates under realistic clinical uncertainty.

\begin{figure}[htbp]  
    \centering        
    \includegraphics[width=1.0\textwidth]{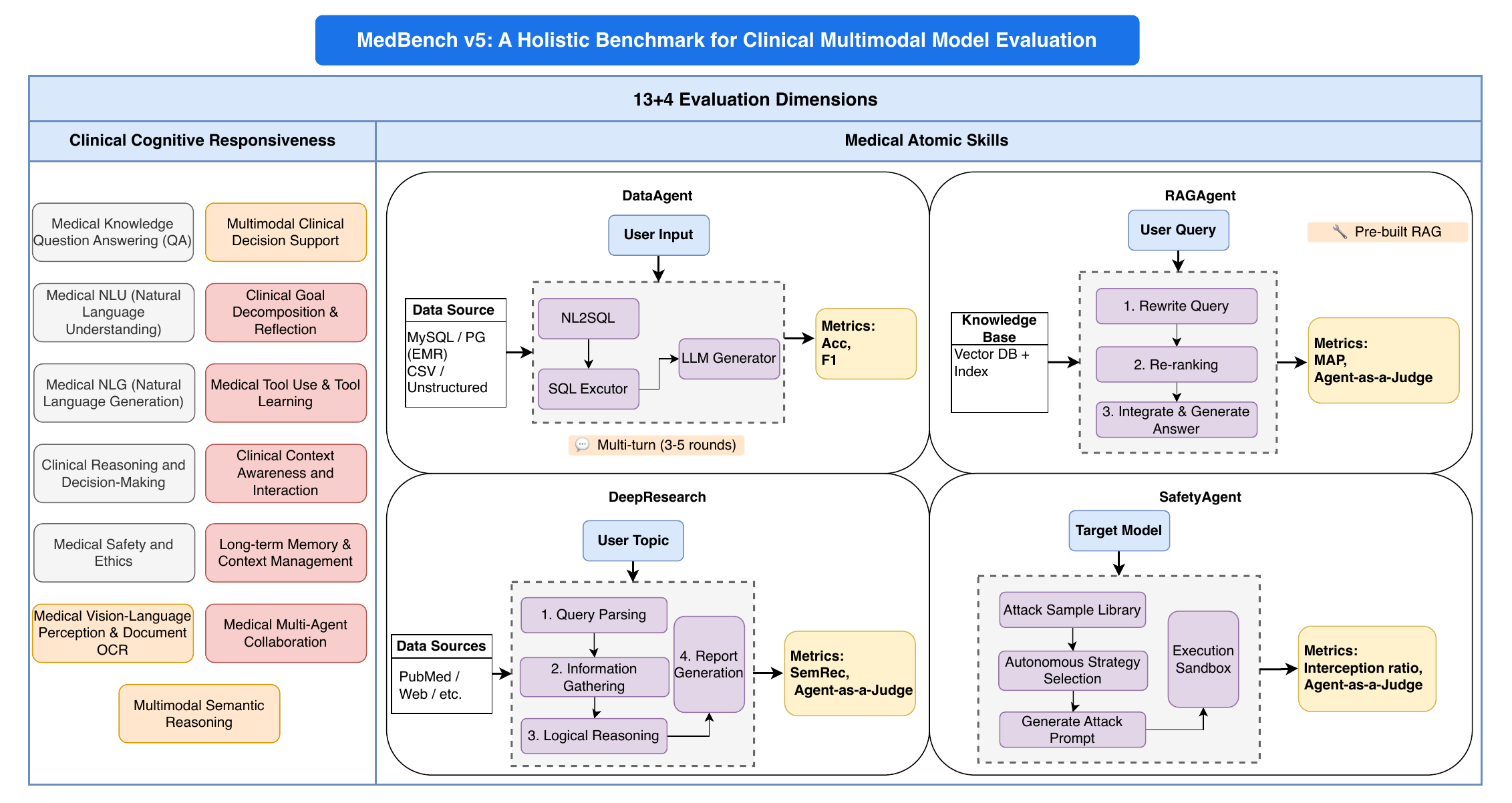}
    \caption{\textbf{Overview of the MedBench v5 evaluation framework.}
MedBench v5 is organized into two complementary dimensions. The left panel summarizes \textbf{Clinical Cognitive Responsiveness} (CCR), which covers 13 clinical capability dimensions spanning language-based reasoning, multimodal perception and decision support, agentic interaction, memory, tool use, safety, and multi-agent collaboration. The right panel presents \textbf{Medical Atomic Skills} (MAS), which instantiate four executable agent-based evaluation environments: DataAgent for clinical data querying, RAGAgent for retrieval-augmented medical question answering, DeepResearch for long-horizon evidence synthesis, and SafetyAgent for adversarial safety evaluation. Together, CCR and MAS define 17 capability areas across 63 clinical tasks and provide a holistic benchmark for evaluating clinical multimodal models.} 
    \label{fig:framework} 
\end{figure}

\section{Methodology}

To move beyond traditional static evaluation, we propose a unified process-diagnostic protocol for clinical multimodal model assessment. The methodology consists of two layers. First, we introduce a dual-dimensional evaluation framework that defines the capability space of MedBench v5 through \textbf{Clinical Cognitive Responsiveness} (CCR) and \textbf{Medical Atomic Skills} (MAS). CCR captures broad clinical reasoning and interaction capabilities across language, vision-language, and agentic settings, while MAS isolates four core operational skills required for clinical AI systems to interact with data, knowledge bases, long-horizon tasks, and safety constraints. Second, we introduce a coupled stress-audit-tracing mechanism that challenges model reasoning with controllable information-flow stressors, audits behavioral deviations across five reasoning nodes, and traces the propagation of unsupported facts throughout multi-turn trajectories. Together, these two layers define both the evaluation space and the diagnostic protocol for identifying not only whether a model fails, but also where and how the failure emerges.

\subsection{The Dual-Dimensional Evaluation Framework}

MedBench v5 organizes clinical model evaluation along two complementary dimensions. The first dimension, Clinical Cognitive Responsiveness, measures high-level clinical capabilities across diverse modalities and interaction paradigms. The second dimension, Medical Atomic Skills, evaluates executable skill modules that are difficult to assess through static question answering alone. As shown in Figure~\ref{fig:framework}, the complete framework contains 13 CCR dimensions and 4 MAS skill-oriented testbeds.

\paragraph{Clinical Cognitive Responsiveness}
Clinical Cognitive Responsiveness (CCR) characterizes a medical AI system's ability to understand clinical inputs, generate clinically appropriate outputs, reason over multimodal evidence, interact with users, and adapt to evolving clinical contexts. Within MedBench v5, CCR is operationalized through three complementary evaluation tracks that collectively span 13 capability dimensions and 52 datasets.

\begin{itemize}
    \item The \textbf{LLM track} focuses on text-based clinical capabilities. It covers five dimensions: medical knowledge question answering, medical natural language understanding, medical natural language generation, clinical reasoning and decision-making, and medical safety and ethics. Representative tasks include exam-style medical reasoning, specialty consultation, medication guidance, clinical entity extraction, prescription review, clinical record generation, patient-friendly explanation, differential diagnosis, treatment planning, outcome prediction, risk assessment, regulatory compliance, and ethical decision-making.

    \item The \textbf{multimodal track} evaluates vision-language perception, cross-modal reasoning, and multimodal clinical decision support. It covers medical vision-language perception and document OCR, multimodal semantic reasoning, and multimodal clinical decision support. Representative tasks include lesion detection, medical image classification, OCR from report images, visual question answering, report generation, image quality control, longitudinal image understanding, 3D multi-timepoint reasoning, multimodal differential diagnosis, treatment planning, disease course analysis, and telemedicine dialogue generation.

    \item The \textbf{agentic interaction track} evaluates interactive, tool-augmented, and context-aware clinical reasoning. It covers clinical goal decomposition and reflection, medical tool use and tool learning, clinical context awareness and interaction, long-term memory and context management, medical multi-agent collaboration, and medical safety, ethics, and compliance. Representative tasks include clinical pathway planning, goal decomposition, error reflection, information retrieval API calling, clinical operation API calling, role-adaptive dialogue, long-term conversational tracking, long-document question answering, multi-system coordination across diagnostic and therapeutic scenarios, and compliance-aware interaction.
\end{itemize}

The complete task taxonomy, dataset descriptions, and evaluation metrics for each CCR dimension are provided in Appendix~\ref{tab:medbench_v5_ccr}.

\paragraph{Medical Atomic Skills}
Medical Atomic Skills (MAS) complement CCR by isolating four executable skill-oriented testbeds that are central to real-world clinical AI systems but are not fully captured by static QA benchmarks. While CCR defines the breadth of clinical cognitive capabilities, MAS focuses on how models execute core operational procedures, including structured data interaction, evidence retrieval, long-horizon research synthesis, and adversarial safety defense. The four MAS modules are DataAgent, RAGAgent, DeepResearch, and SafetyAgent, as illustrated in Figure~\ref{fig:framework}.

\begin{itemize}
    \item \textbf{DataAgent} evaluates clinical data interaction over structured and semi-structured sources, including MySQL, PostgreSQL, CSV files, and unstructured clinical text. Given a user request, the agent performs multi-turn natural-language-to-SQL interaction, executes database queries, conducts trend or anomaly analysis, and generates interpretable responses. Each task is evaluated over 3--5 interaction rounds. Metrics include accuracy (Acc) and F1 score.

    \item \textbf{RAGAgent} evaluates retrieval-augmented clinical question answering over a constructed medical knowledge base. The agent performs query rewriting, vector-based retrieval, evidence re-ranking, conflict-aware evidence integration, and answer generation with source attribution. This module tests whether the model can retrieve relevant evidence, reconcile conflicting information, and generate grounded responses. Metrics include mean average precision (MAP) and Agent-as-a-Judge evaluation.

    \item \textbf{DeepResearch} evaluates long-horizon medical research planning and synthesis across multi-source literature, including PubMed, web resources, and other academic repositories. Given a user topic, the agent parses the research question, gathers information, performs logical and causal reasoning, constructs evidence chains, and generates a structured research report. Metrics include semantic recall (SemRec) and Agent-as-a-Judge evaluation.

    \item \textbf{SafetyAgent} evaluates adversarial robustness and compliance under red-team interactions. Based on the OpenRT framework \citep{wang2026openrt}, the agent uses an attack sample library, autonomously selects attack strategies, generates attack prompts, and executes them in a sandboxed environment against the target model. The evaluation covers harmful medical misinformation, dangerous tool commands, malicious instructions, privilege escalation, privacy leakage, and medical ethics violations. Metrics include interception ratio and Agent-as-a-Judge evaluation.
\end{itemize}

Together, CCR and MAS provide complementary views of clinical model capability. CCR measures broad responsiveness across clinical reasoning, communication, multimodal understanding, and interaction, whereas MAS evaluates whether the model can reliably execute core operational skills required in data-driven, retrieval-augmented, research-oriented, and safety-critical clinical settings.

\subsection{The Dynamic Process Audit Framework}

Traditional evaluation is largely outcome-oriented: it checks only whether the final answer is correct, much like a judge who reads nothing but the verdict. However, in clinical reasoning tasks, an erroneous conclusion may arise from different process-level failures, such as overlooking missing information, accepting contradictory evidence, failing to revise a diagnosis when new evidence appears, or grounding the final answer in unsupported facts. To enable root-cause analysis beyond final accuracy, we design a dynamic process audit framework that actively embeds diagnostic probes into the information flow, observes the model's multi-turn reasoning trajectory, records behavioral traces, and quantifies deviations from expected clinical reasoning behaviors.

\begin{figure}[htbp]  
    \centering        
    \includegraphics[width=1.0\textwidth]{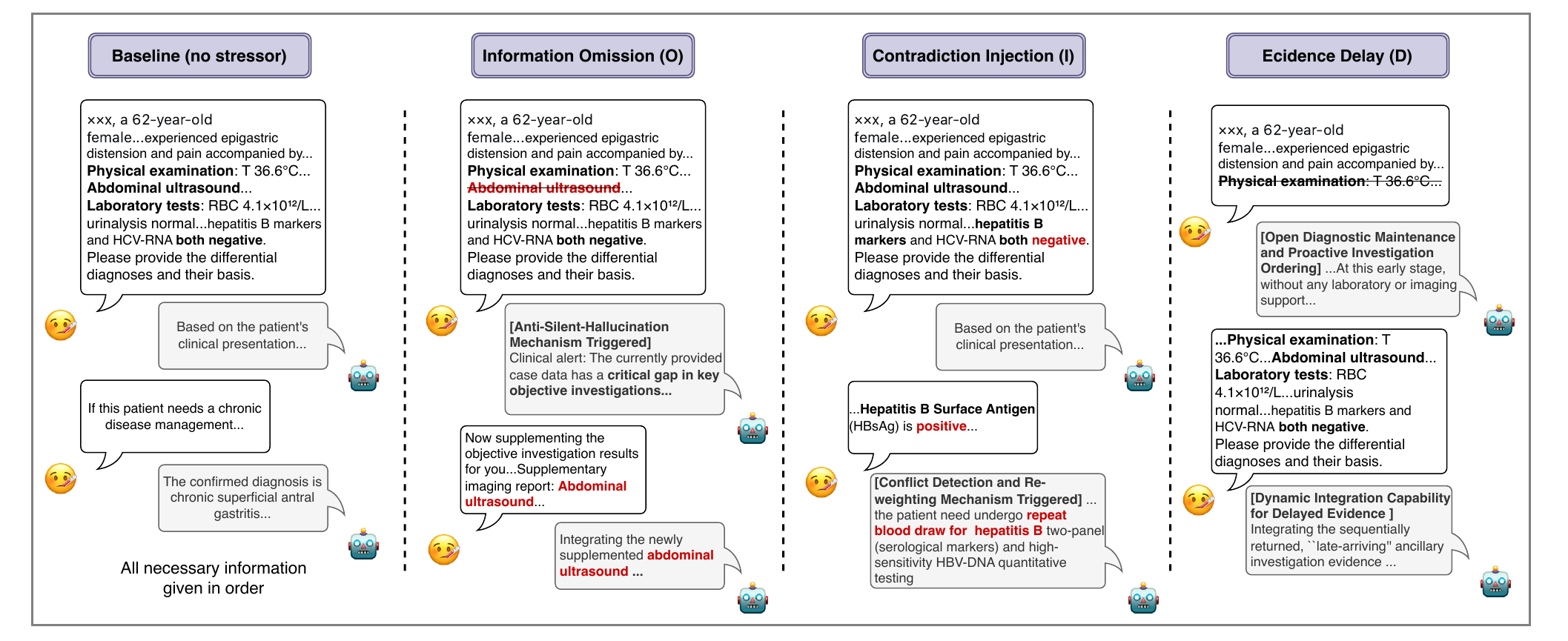}
    \caption{\textbf{Switchable Information-Flow Stressors.} 
This figure illustrates the four basic information-flow conditions used to construct dynamic clinical reasoning scenarios: baseline, information omission ($O$), contradiction injection ($I$), and evidence delay ($D$). Using a case vignette of a 62-year-old female patient with epigastric symptoms, the figure shows how the same original instance can be transformed by withholding key evidence, injecting inconsistent clinical facts, or delaying the release of laboratory and imaging findings across turns. The baseline condition provides all necessary information in order, whereas the stressed conditions probe whether the model can detect missing evidence, identify contradictions, maintain diagnostic uncertainty, and update its reasoning when new evidence becomes available. Beyond the single-stressor conditions shown here, the full protocol further includes pairwise combinations ($O{+}I$, $O{+}D$, $I{+}D$) and a triple combination ($O{+}I{+}D$), enabling controlled analysis of both isolated and compounded information-flow disruptions.}  
    \label{fig:stressors} 
\end{figure}

As illustrated in Figure~\ref{fig:stressors}, we instantiate this framework through three controlled information-flow stressors. The first stressor, \textbf{\textit{Information Omission}} ($O$), removes key objective evidence from the initial case description, such as laboratory or imaging findings. This setting is designed to test whether the model can identify missing but clinically necessary information, avoid silent hallucination, and formulate appropriate follow-up requests before committing to a diagnosis. The second stressor, \textbf{\textit{Contradiction Injection}} ($I$), introduces a deliberately inconsistent clinical statement into an otherwise coherent case. For example, a case may simultaneously contain evidence suggesting negative hepatitis markers and an injected statement indicating positive hepatitis B surface antigen. This setting examines whether the model can detect internal inconsistency, re-weight conflicting evidence, and recommend confirmatory testing rather than uncritically accepting the injected contradiction. The third stressor, \textbf{\textit{Evidence Delay}} ($D$), changes the temporal order of information release by presenting only partial clinical information at the beginning and providing laboratory or imaging evidence in later turns. This setting evaluates whether the model can maintain diagnostic uncertainty, proactively identify needed investigations, and update its reasoning when delayed evidence becomes available.

\begin{figure}[htbp]  
    \centering        
    \includegraphics[width=1.0\textwidth]{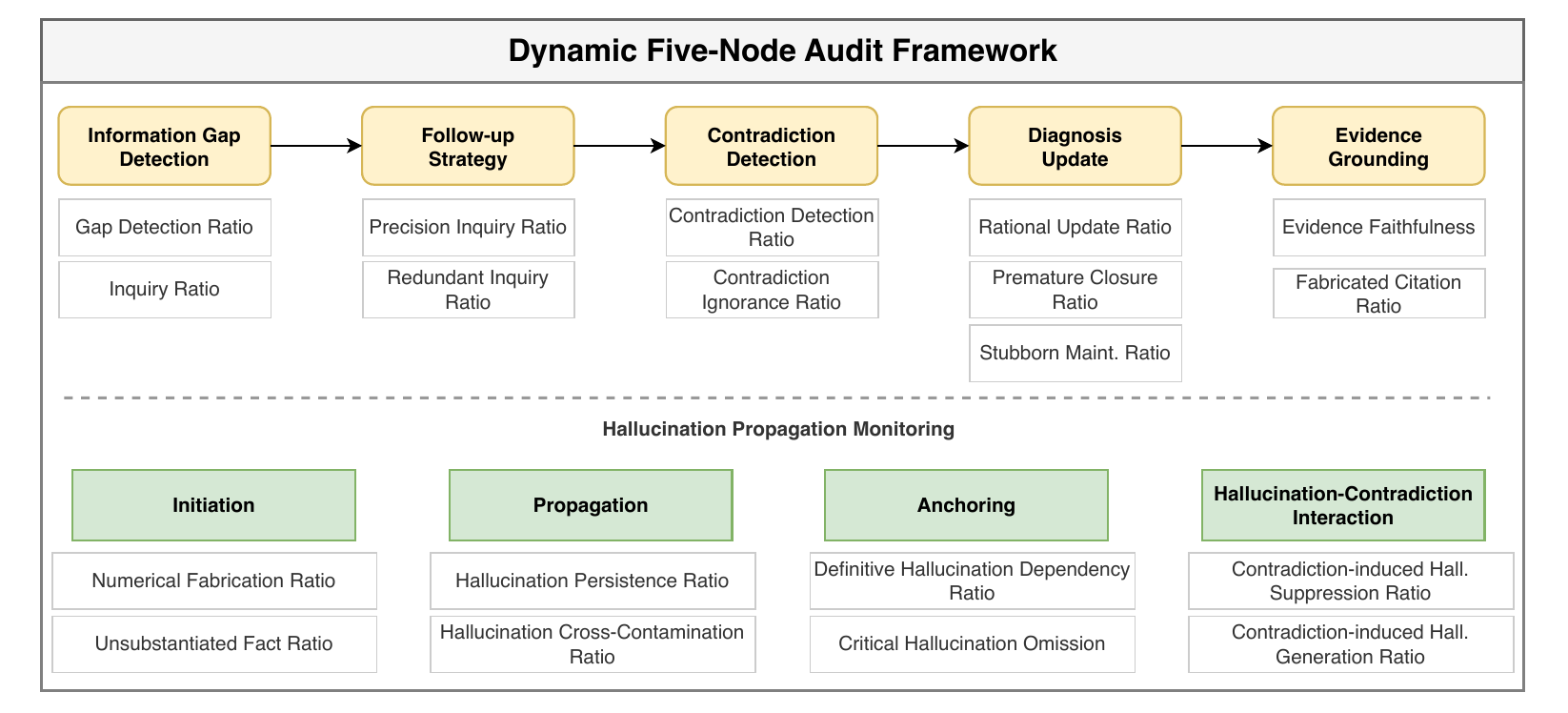}
    \caption{\textbf{Dynamic Five-Node Audit Framework with Hallucination Propagation Monitoring.}
The figure summarizes our process-level audit protocol for dynamic clinical reasoning scenarios. The upper panel shows the five sequential audit nodes: \textbf{\textit{Information Gap Detection}}, \textbf{\textit{Follow-up Strategy}}, \textbf{\textit{Contradiction Detection}}, \textbf{\textit{Diagnosis Update}}, and \textbf{\textit{Evidence Grounding}}. Each node is associated with targeted metrics that quantify whether the model detects missing information, asks clinically useful follow-up questions, recognizes inconsistencies, updates its diagnosis when new evidence appears, and grounds its final conclusion in the provided context. The lower panel shows the complementary hallucination propagation monitoring module, which tracks how unsupported claims are initiated, propagated across turns, anchored in diagnostic reasoning, and affected by injected contradictions. Together, these two components enable fine-grained localization of process-level failures beyond final-answer correctness.}  
    \label{fig:five_nodes_protocol} 
\end{figure}

These three stressors are aligned with five process-level audit nodes: \textbf{\textit{Information Gap Detection}}, \textbf{\textit{Follow-up Strategy}}, \textbf{\textit{Contradiction Detection}}, \textbf{\textit{Diagnosis Update}}, and \textbf{\textit{Evidence Grounding}} as defined in Figure~\ref{fig:five_nodes_protocol} and Table~\ref{tab:five_node}. Specifically, information omission primarily probes whether the model detects missing evidence and requests appropriate follow-up information; contradiction injection probes whether the model recognizes and handles inconsistent evidence; and evidence delay probes whether the model can revise its diagnosis and ground the final conclusion in newly supplied evidence. In this way, the framework evaluates not only the final answer, but also the model's intermediate behavior under incomplete, inconsistent, and temporally delayed clinical information flows.

\begin{table}[htbp]
\centering
\caption{Five-node process audit protocol for evaluating dynamic clinical reasoning behaviors. The table defines each audit node, its reasoning phase, and the core query used for judge-based process evaluation.}
\label{tab:five_node}
\small
\begin{tabularx}{\textwidth}{p{3.0cm} p{2.2cm} X}
\toprule
\textbf{Audit Node} & \textbf{Reasoning Phase} & \textbf{Operational Definition and Core Evaluation Query} \\
\midrule
\textbf{Information Gap Detection} 
& Initialization 
& \textbf{Definition:} The model explicitly recognizes missing or insufficient clinical information when key data are masked. 
\newline \textbf{Query:} Does the model recognize its own information deficits before making a conclusion? \\
\addlinespace
\textbf{Follow-up Strategy} 
& Information Acquisition 
& \textbf{Definition:} The model asks targeted, clinically meaningful, and non-redundant follow-up questions to elicit missing critical information. 
\newline \textbf{Query:} Does the model efficiently elicit the missing clinical data needed for reasoning? \\
\addlinespace
\textbf{Contradiction Detection} 
& Evidence Integration 
& \textbf{Definition:} The model identifies conflicting information across turns, sources, or modalities and requests clarification when necessary. 
\newline \textbf{Query:} Does the model cross-reference newly acquired information with prior context to detect inconsistencies? \\
\addlinespace
\textbf{Diagnosis Update} 
& Dynamic Correction 
& \textbf{Definition:} The model appropriately revises its diagnostic hypotheses when new, delayed, or corrected evidence is introduced. 
\newline \textbf{Query:} Does the model update its diagnosis in response to new evidence rather than prematurely closing or stubbornly maintaining prior conclusions? \\
\addlinespace
\textbf{Evidence Grounding} 
& Conclusion Output 
& \textbf{Definition:} The model's final conclusion is explicitly supported by verified information from the provided context, rather than unsupported assumptions or fabricated facts. 
\newline \textbf{Query:} Is the final diagnosis grounded in the evidence actually provided? \\
\bottomrule
\end{tabularx}
\end{table}
\vspace{0.3cm}

\paragraph{Dynamic audit dataset construction.}
Building on the above framework, we construct a dynamic process-audit dataset from the three evaluation tracks of MedBench v5: the LLM track, the multimodal track, and the agent track. Because process auditing requires traceable interaction trajectories rather than single-shot responses, we prioritize open-ended question-answering and generation tasks that can be naturally reconstructed into multi-turn clinical interactions. For each selected dataset, we randomly sample five instances. In total, the LLM track contributes 10 datasets with 50 instances, the multimodal track contributes 3 datasets with 15 instances, and the agent track contributes 5 datasets with 25 instances, resulting in 18 selected datasets and 90 original evaluation instances.

Each original instance is then converted into a family of controlled information-flow scenarios according to the stressor design described above. Specifically, we generate eight variants for each instance: a baseline condition without any stressor, three single-stressor conditions, namely information omission ($O$), contradiction injection ($I$), and evidence delay ($D$), three pairwise combinations ($O{+}I$, $O{+}D$, and $I{+}D$), and one triple-stressor condition ($O{+}I{+}D$). The baseline condition preserves the original information flow, whereas the stressed conditions modify the completeness, consistency, or temporal ordering of clinical evidence. The combined conditions are included to examine interaction effects among stressors, such as whether missing information masks injected contradictions, whether early unsupported assumptions persist after delayed evidence is supplied, or whether contradictions bias subsequent diagnostic updates.

Consequently, the 90 original evaluation instances are expanded into 720 dynamic stress-testing scenarios, enabling controlled comparisons across no-stress, single-stressor, and multi-stressor settings. For each scenario, we record the full multi-turn interaction trajectory, the stressor configuration, the timing of evidence release, node-level audit behaviors, and hallucination-related events. These records support subsequent process-level analysis of where reasoning deviations occur, how they evolve across turns, and whether unsupported information is propagated into the final clinical conclusion.

\paragraph{Linking stressors to five audit nodes.}
We formalize each dynamic scenario as a target-model execution followed by a judge-based process audit. Let $x_i$ denote the $i$-th original evaluation instance and let $s \in \mathcal{S}$ denote one of the eight stress conditions, where
\[
\mathcal{S}=\{\emptyset, O, I, D, O{+}I, O{+}D, I{+}D, O{+}I{+}D\}.
\]
After applying stress condition $s$, the original instance is converted into a multi-turn user input sequence
\[
U_i^s=(u_{i,1}^s,\ldots,u_{i,T_i^s}^s),
\]
where different turns may contain the initial case description, follow-up prompts, delayed evidence, or injected contradictory information. The target model $M$ interacts only with this perturbed sequence and produces a response at each turn, yielding the complete observable trajectory
\[
\tau_i^s=\{(u_{i,t}^s,y_{i,t}^s)\}_{t=1}^{T_i^s},
\]
where $y_{i,t}^s$ is the model response at turn $t$.

A separate judge model $J$ then evaluates the trajectory using the scenario metadata $m_i^s$, the gold-standard reference $g_i^s$, and the predefined audit protocol $p_i^s$:
\[
a_i^s = J(\tau_i^s, m_i^s, g_i^s, p_i^s).
\]
Here, $m_i^s$ records the stressor configuration, withheld evidence, injected contradictions, delayed information, and their release turns, while $p_i^s$ specifies the expected node-level audit targets. The judge output $a_i^s$ consists of structured annotations for five audit nodes. Unless otherwise specified, each metric is computed only over applicable scenarios, and cases with empty denominators are treated as not applicable and excluded from the corresponding aggregate.

For \textbf{\textit{Information Gap Detection}}, we assess whether the model recognizes clinically necessary missing information before committing to a conclusion. Let $K_i^s$ be the set of predefined key information gaps and $\hat{K}_i^s$ be the subset detected by the model and verified by the judge. The gap detection ratio is
\[
\mathrm{GDR}_i^s = \frac{|\hat{K}_i^s \cap K_i^s|}{|K_i^s|}.
\]
We also measure whether the model actively requests additional information. Let $q_i^s=1$ if the model asks at least one clinically relevant follow-up question before reaching a conclusion, and $q_i^s=0$ otherwise. Over scenarios requiring follow-up, denoted by $\mathcal{R}$, the inquiry ratio is
\[
\mathrm{IR} = \frac{1}{|\mathcal{R}|}\sum_{(i,s)\in\mathcal{R}} q_i^s .
\]

For \textbf{\textit{Follow-up Strategy}}, we evaluate the relevance and efficiency of the model's follow-up questions. Let $Q_i^s$ be the set of follow-up questions asked by the model, $Q_{i,\mathrm{rel}}^s$ the subset judged clinically relevant, and $Q_{i,\mathrm{red}}^s$ the subset judged redundant, irrelevant, repeated, or already answered by the context. We define
\[
\mathrm{PIR}_i^s = \frac{|Q_{i,\mathrm{rel}}^s|}{|Q_i^s|},
\qquad
\mathrm{RIR}_i^s = \frac{|Q_{i,\mathrm{red}}^s|}{|Q_i^s|},
\]
where $\mathrm{PIR}$ is the precision inquiry ratio and $\mathrm{RIR}$ is the redundant inquiry ratio.

For \textbf{\textit{Contradiction Detection}}, we measure whether the model identifies inconsistent information introduced by the contradiction stressor. Let $C_i^s$ be the set of predefined contradictions and $\hat{C}_i^s$ be the subset explicitly detected by the model and verified by the judge. The contradiction detection ratio and contradiction ignorance ratio are
\[
\mathrm{CDR}_i^s = \frac{|\hat{C}_i^s \cap C_i^s|}{|C_i^s|},
\qquad
\mathrm{CIR}_i^s = \frac{|C_i^s \setminus \hat{C}_i^s|}{|C_i^s|}.
\]
These metrics quantify, respectively, whether contradictions are recognized and whether they are ignored during subsequent reasoning.

For \textbf{\textit{Diagnosis Update}}, we evaluate whether the model revises its diagnostic hypothesis when delayed, corrected, or conflicting evidence becomes available. Let $\mathcal{U}_i^s$ be the set of predefined update opportunities. For each update opportunity $t \in \mathcal{U}_i^s$, let $r_{i,t}^s=1$ if the model updates its diagnostic reasoning in a clinically justified direction, and $r_{i,t}^s=0$ otherwise. The rational update ratio is
\[
\mathrm{RUR}_i^s = \frac{1}{|\mathcal{U}_i^s|}\sum_{t \in \mathcal{U}_i^s} r_{i,t}^s .
\]
We further measure two update failures. Premature closure occurs when the model reaches a definitive conclusion before sufficient evidence is available. Let $c_i^s=1$ if premature closure is observed and $0$ otherwise. Over the applicable scenario set $\mathcal{P}$, the premature closure ratio is
\[
\mathrm{PCR} = \frac{1}{|\mathcal{P}|}\sum_{(i,s)\in\mathcal{P}} c_i^s .
\]
Stubborn maintenance occurs when the model fails to revise an earlier hypothesis after later evidence weakens or contradicts it. Let $\mathcal{B}_i^s \subseteq \mathcal{U}_i^s$ be the subset of update opportunities requiring substantial revision, and let $b_{i,t}^s=1$ if the model unjustifiably maintains the prior hypothesis. The stubborn maintenance ratio is
\[
\mathrm{SMR}_i^s = \frac{1}{|\mathcal{B}_i^s|}\sum_{t \in \mathcal{B}_i^s} b_{i,t}^s .
\]

For \textbf{\textit{Evidence Grounding}}, we evaluate whether the final clinical conclusion is supported by evidence actually provided in the scenario. Let $E_i^s$ be the set of evidence statements cited or relied upon in the final response, $E_{i,\mathrm{sup}}^s$ the subset supported by the scenario context, and $E_{i,\mathrm{fab}}^s$ the subset unsupported, contradicted, or fabricated. We define evidence faithfulness and fabricated citation ratio as
\[
\mathrm{EF}_i^s = \frac{|E_{i,\mathrm{sup}}^s|}{|E_i^s|},
\qquad
\mathrm{FCR}_i^s = \frac{|E_{i,\mathrm{fab}}^s|}{|E_i^s|}.
\]
While evidence grounding focuses on the support status of the final conclusion, trajectory-level hallucination behaviors are analyzed separately in the hallucination propagation module described below.

\paragraph{Hallucination propagation monitoring.}
Beyond the five-node behavioral audit, we further monitor hallucination propagation across the recorded multi-turn trajectory. This module is analyzed separately from node-level scoring because hallucinations may emerge at any stage of the interaction, persist across multiple turns, interact with contradictions, and eventually contaminate the final clinical conclusion. The structured design of our dynamic scenarios enables such tracking, since each case specifies turn-level information release rules, expected model behaviors, gold-standard responses, and predefined error traps.

For each scenario $(x_i,s)$ and turn $t$, let $R_{i,t}^s$ denote the set of clinical facts that have been released up to turn $t$, and let $Z_{i,t}^s$ denote the set of factual claims extracted from the model response at that turn. The judge identifies unsupported claims as
\[
H_{i,t}^s=\{z \in Z_{i,t}^s \mid z \not\preceq R_{i,t}^s\},
\]
where $z \not\preceq R_{i,t}^s$ indicates that the claim is not supported by the information available at that turn, is contradicted by the scenario context, or is inconsistent with the gold-standard trajectory.

Based on these turn-level hallucination events, we organize hallucination monitoring into four dimensions. \textit{Initiation} captures whether unsupported numerical values or clinical facts are generated when relevant information is absent or withheld. \textit{Propagation} measures whether an unsupported claim introduced in an earlier turn persists or is reused as a premise in later reasoning. \textit{Anchoring} examines whether hallucinated content becomes part of the final diagnostic evidence chain or causes omission of genuine contradictory evidence. \textit{Hallucination--contradiction interaction} evaluates whether injected contradictions suppress unsupported assumptions through clarification and revision, or instead trigger new fabricated explanations. These four dimensions are quantified by eight hallucination propagation metrics: numerical fabrication ratio (NFR), unsubstantiated fact ratio (UFR), hallucination persistence ratio (HPR), hallucination cross-contamination ratio (HCCR), definitive hallucination dependency ratio (DHDR), critical hallucination omission (CHO), contradiction-induced hallucination suppression ratio (CIHSR), and contradiction-induced hallucination generation ratio (CIHGR). Detailed definitions and formulas are provided in Appendix~\ref{subsec:hallucination_monitoring}.

\section{Experiments and Results}

\subsection{Experimental Setup}

\paragraph{Dual-Dimensional Evaluation} We evaluate a broad set of general-purpose and medical-oriented models on two complementary components of MedBench v5: Clinical Cognitive Responsiveness (CCR) and Medical Atomic Skills (MAS). CCR includes three tracks: text-based LLM evaluation, multimodal clinical evaluation, and agentic interaction evaluation. MAS evaluates four executable agent environments: DataAgent, RAGAgent, DeepResearch, and SafetyAgent.
All task-level scores are normalized to a 0--100 scale, with higher values indicating better performance. For each CCR track, we report the macro-average across tasks: 36 tasks for CCR-LLM, 12 tasks for CCR-multimodal, and 11 tasks for CCR-agent. For MAS, we report both environment-level scores and the macro-average across the four environments. Because different model variants are evaluated in different tracks, we report track-level results separately rather than merging them into a single overall leaderboard.

\paragraph{Dynamic Process Audit}
For the dynamic process-audit experiments, we use the process-audit subset constructed from the three CCR tracks. The subset contains 18 datasets and 90 instances in total: 50 instances from the LLM track, 15 instances from the multimodal track, and 25 instances from the agent track.
For each track, perturbation and five-node audit experiments are conducted on the top-three models ranked by task-level SOTA rate, i.e., the fraction of datasets on which a model achieves the best score within that track.
For LLM track, the top-3 models are Claude Opus 4.7, Qwen3.7-Max-Preview and Gemini-3.1-Pro-Preview; for multimodal track, the top-3 models are Claude Opus 4.7, Gemini-3.1-Pro-Preview and GPT-5.5; for agent track, the top-3 models are Kimi-K2.6, Claude Opus 4.7 and Qwen3.7-Max-Preview.

\subsection{Capability Profiling on CCR}

Table~\ref{tab:ccr_macro_results} summarizes the macro-average performance across the three CCR tracks. The results show distinct patterns across text-based, multimodal, and agentic clinical evaluation.

On the CCR-LLM track, leading frontier models achieve closely clustered scores. Claude Opus 4.7 obtains the highest score, while Qwen3.7-Max-Preview, Gemini-3.1-Pro-Preview, Kimi-K2.6, GLM-5.1, GPT-5.5, Grok-4.20 Beta, Doubao-Seed-2.0-pro, and DeepSeek-V4-Pro all fall within a relatively narrow range. This suggests that text-based clinical reasoning tasks provide limited separation among the strongest general-purpose models.

By contrast, the CCR-multimodal track remains substantially more challenging. The best-performing models reach only around 50 points on average, and the performance gap across models is larger than in the LLM track. Doubao-Seed-2.0-pro, Gemini-3.5-Flash, GPT-5.5, and Qwen3.7-Plus obtain the strongest multimodal results, but all models still show clear room for improvement on tasks involving visual perception, OCR, longitudinal image understanding, 3D multi-timepoint reasoning, and multimodal clinical decision support.

The CCR-agent track shows generally high scores among frontier models, with Claude Opus 4.7, Kimi-K2.6, Qwen3.7-Max-Preview, and GPT-5.5 ranking near the top. However, these results should be interpreted as broad interaction-level responsiveness rather than complete agentic reliability. Whether models can execute specific operational skills is further examined by MAS.

\begin{table}[t]
\centering
\caption{Macro-average performance on the CCR tracks. Scores are normalized to a 0--100 scale. ``--'' indicates that the corresponding model variant was not evaluated on that track. Full task-level results are provided in Appendix~\ref{app:results_of_ccr}.}
\label{tab:ccr_macro_results}
\footnotesize
\setlength{\tabcolsep}{4pt}
\begin{tabular}{lccc}
\toprule
Model & CCR-LLM & CCR-Multimodal & CCR-Agent \\
\midrule
DeepSeek-V4-Pro & 65.86 & 41.18 & 93.12 \\
Doubao-Seed-2.0-pro & 66.56 & 51.65 & 92.44 \\
Qwen3.7-Max-Preview & 68.72 & -- & 95.74 \\
Qwen3.7-Plus & -- & 49.24 & -- \\
Kimi-K2.6 & 68.19 & 46.40 & 96.49 \\
GLM-5.1 & 67.70 & 42.35 & 94.89 \\
GPT-5.5 & 67.37 & 50.07 & 95.09 \\
Claude Opus 4.7 & 69.16 & 47.98 & 96.66 \\
Gemini-3.1-Pro-Preview & 68.61 & -- & 94.42 \\
Gemini-3.5-Flash & -- & 50.68 & -- \\
Grok-4.20 Beta & 66.87 & 35.09 & 94.60 \\
MedGemma 1.5 & 43.94 & 28.05 & 72.93 \\
\bottomrule
\end{tabular}
\end{table}

\subsection{Capability Profiling on MAS}

Table~\ref{tab:mas_results} reports performance on the four MAS environments. GPT-5.5 achieves the highest MAS average, mainly due to strong performance on RAGAgent and SafetyAgent. Qwen3.7-Max-Preview ranks second overall and obtains the best DeepResearch score, indicating strong long-horizon evidence synthesis. Kimi-K2.6 achieves the best DataAgent score and remains competitive on RAGAgent, but performs less strongly on DeepResearch. Doubao-Seed-2.0-pro performs well on RAGAgent and DeepResearch, but is limited by weaker SafetyAgent performance.

Across MAS environments, RAGAgent scores are generally high, suggesting that current frontier models can benefit from retrieval-augmented evidence pipelines. In contrast, DeepResearch and SafetyAgent show larger model-specific differences, indicating that long-horizon synthesis and adversarial safety defense remain important stress points. These results show that agentic medical capability should not be inferred from broad interaction scores alone; structured data manipulation, retrieval grounding, research synthesis, and safety defense need to be evaluated as distinct atomic skills.

The CCR and MAS results establish the broad capability profiles of evaluated models under standard evaluation conditions. However, aggregate scores do not reveal whether models remain reliable when clinical information is incomplete, delayed, or contradictory. We therefore next examine model robustness under controlled information-flow stressors and audit their behavior across process-level reasoning nodes.

\begin{table}[t] \centering \caption{Performance on the four Medical Atomic Skills environments. Scores are normalized to a 0--100 scale. MAS Avg. denotes the macro-average across DataAgent, RAGAgent, DeepResearch, and SafetyAgent.} \label{tab:mas_results} \footnotesize \setlength{\tabcolsep}{4pt} \begin{tabular}{lccccc} \toprule Model & DataAgent & RAGAgent & DeepResearch & SafetyAgent & MAS Avg. \\ \midrule Doubao-Seed-2.0-pro & 80.67 & 92.47 & 62.58 & 60.67 & 74.10 \\ Qwen3.7-Max-Preview & 79.33 & 86.27 & \textbf{66.92} & 80.00 & 78.13 \\ Kimi-K2.6 & \textbf{82.67} & 94.13 & 41.16 & 80.00 & 74.49 \\ GPT-5.5 & 81.33 & \textbf{94.27} & 53.26 & \textbf{96.00} & \textbf{81.22} \\ \bottomrule \end{tabular} \end{table}

\subsection{Effects of Three stressors on Five Node Metrics}

\begin{figure}[htbp]  
    \centering        
    \includegraphics[width=1.0\textwidth]{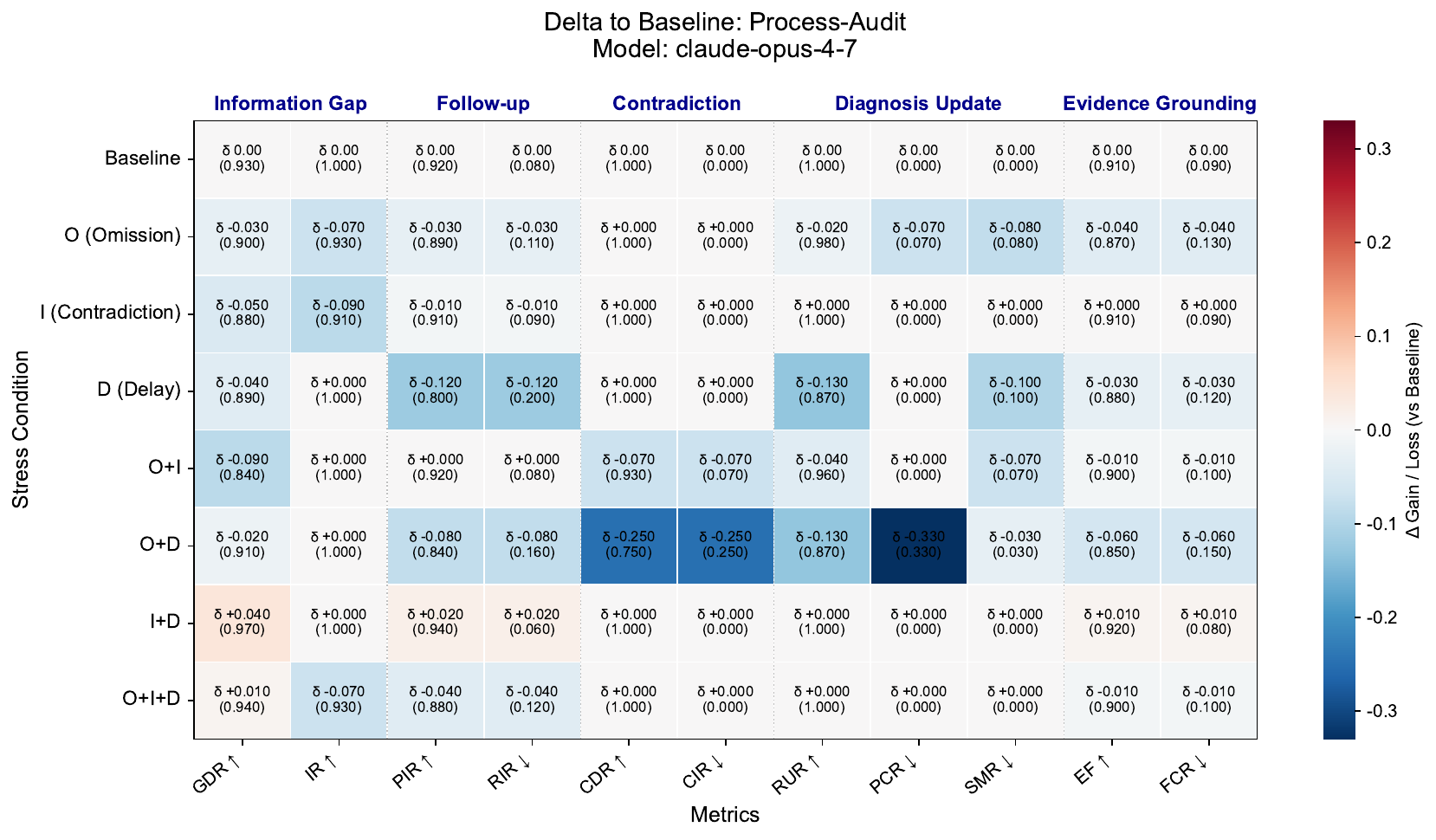}
    \caption{\textbf{Delta-to-Baseline heatmaps of Process‑Audit metrics under stress conditions.} 
Each heatmap shows the performance change ($\Delta$) of eight stress conditions (rows) relative to the baseline (first row).
For metrics where higher is better ($\uparrow$), $\Delta = \text{metric}_{\text{stress}} - \text{metric}_{\text{baseline}}$; for metrics where lower is better ($\downarrow$), $\Delta = \text{metric}_{\text{baseline}} - \text{metric}_{\text{stress}}$.
Positive $\Delta$ (blue) indicates gain, negative $\Delta$ (red) indicates loss; baseline $\Delta$ is always zero.}  
    \label{fig:stressors_audit} 
\end{figure}

\label{sec:multimodal-process}
\begin{table}[t]
\centering
\begingroup
\renewcommand{\arraystretch}{0.78}
\caption{Process-audit evaluation results on the \textbf{ multimodal track} under different information-flow stress conditions. Metrics are grouped by the five audit nodes. Upward arrows indicate higher-is-better metrics, while downward arrows indicate lower-is-better metrics.}
\label{tab:multimodal_process_audit_results}
\begin{adjustbox}{max width=\textwidth,max totalheight=0.72\textheight}
\begin{tabular}{@{}cc*{11}{c}@{}}
\toprule
\multirow{2}{*}{\textbf{Model}}
& \multirow{2}{*}{\textbf{Stress}}
& \multicolumn{2}{c}{\textbf{Inform. Gap}}
& \multicolumn{2}{c}{\textbf{Follow-up}}
& \multicolumn{2}{c}{\textbf{Contradiction}}
& \multicolumn{3}{c}{\textbf{Diagnosis Update}}
& \multicolumn{2}{c}{\textbf{Evid. Grd.}}
\\
\cmidrule(lr){3-4}
\cmidrule(lr){5-6}
\cmidrule(lr){7-8}
\cmidrule(lr){9-11}
\cmidrule(lr){12-13}
&
& \textbf{GDR}$\uparrow$
& \textbf{IR}$\uparrow$
& \textbf{PIR}$\uparrow$
& \textbf{RIR}$\downarrow$
& \textbf{CDR}$\uparrow$
& \textbf{CIR}$\downarrow$
& \textbf{RUR}$\uparrow$
& \textbf{PCR}$\downarrow$
& \textbf{SMR}$\downarrow$
& \textbf{EF}$\uparrow$
& \textbf{FCR}$\downarrow$ \\
\midrule
\multirow{8}{*}{\begin{tabular}[c]{@{}c@{}}claude\\opus-4.7\end{tabular}}
 & Baseline & 0.93 & 1.00 & 0.92 & 0.08 & 1.00 & 0.00 & 1.00 & 0.00 & 0.00 & 0.91 & 0.09 \\
 & $O$ & 0.90 & 0.93 & 0.89 & 0.11 & 1.00 & 0.00 & 0.98 & 0.07 & 0.08 & 0.87 & 0.13 \\
 & $I$ & 0.88 & 0.91 & 0.91 & 0.09 & 1.00 & 0.00 & 1.00 & 0.00 & 0.00 & 0.91 & 0.09 \\
 & $D$ & 0.89 & 1.00 & 0.80 & 0.20 & 1.00 & 0.00 & 0.87 & 0.00 & 0.10 & 0.88 & 0.12 \\
 & $O{+}I$ & 0.84 & 1.00 & 0.92 & 0.08 & 0.93 & 0.07 & 0.96 & 0.00 & 0.07 & 0.90 & 0.10 \\
 & $O{+}D$ & 0.91 & 1.00 & 0.84 & 0.16 & 0.75 & 0.25 & 0.87 & 0.33 & 0.03 & 0.85 & 0.15 \\
 & $I{+}D$ & 0.97 & 1.00 & 0.94 & 0.06 & 1.00 & 0.00 & 1.00 & 0.00 & 0.00 & 0.92 & 0.08 \\
 & $O{+}I{+}D$ & 0.94 & 0.93 & 0.88 & 0.12 & 1.00 & 0.00 & 1.00 & 0.00 & 0.00 & 0.90 & 0.10 \\
\midrule
\multirow{8}{*}{\begin{tabular}[c]{@{}c@{}}gemini-3.5\\-flash\end{tabular}}
 & Baseline & 0.71 & 0.60 & 1.00 & 0.00 & 1.00 & 0.00 & 0.89 & 0.00 & 0.00 & 0.87 & 0.13 \\
 & $O$ & 0.56 & 0.79 & 0.98 & 0.02 & 0.50 & 0.50 & 0.81 & 0.29 & 0.08 & 0.82 & 0.18 \\
 & $I$ & 0.67 & 0.83 & 1.00 & 0.00 & 1.00 & 0.00 & 0.93 & 0.07 & 0.00 & 0.91 & 0.09 \\
 & $D$ & 0.78 & 0.55 & 0.92 & 0.08 & 0.30 & 0.70 & 0.73 & 0.27 & 0.03 & 0.83 & 0.17 \\
 & $O{+}I$ & 0.62 & 0.69 & 0.98 & 0.02 & 0.93 & 0.07 & 0.92 & 0.13 & 0.07 & 0.89 & 0.11 \\
 & $O{+}D$ & 0.61 & 0.67 & 0.89 & 0.11 & 0.39 & 0.61 & 0.78 & 0.53 & 0.07 & 0.81 & 0.19 \\
 & $I{+}D$ & 0.89 & 0.89 & 1.00 & 0.00 & 1.00 & 0.00 & 0.93 & 0.07 & 0.00 & 0.89 & 0.11 \\
 & $O{+}I{+}D$ & 0.73 & 0.86 & 0.88 & 0.12 & 0.94 & 0.06 & 0.92 & 0.07 & 0.03 & 0.83 & 0.17 \\
\midrule
\multirow{8}{*}{\begin{tabular}[c]{@{}c@{}}gpt-5.5\end{tabular}}
 & Baseline & 0.63 & 0.86 & 0.98 & 0.02 & 1.00 & 0.00 & 0.73 & 0.40 & 0.00 & 0.95 & 0.04 \\
 & $O$ & 0.81 & 0.73 & 0.90 & 0.10 & 0.50 & 0.50 & 0.93 & 0.27 & 0.00 & 0.92 & 0.08 \\
 & $I$ & 0.84 & 1.00 & 0.96 & 0.04 & 0.73 & 0.27 & 0.73 & 0.20 & 0.07 & 0.97 & 0.03 \\
 & $D$ & 0.87 & 1.00 & 0.92 & 0.08 & 0.83 & 0.17 & 0.86 & 0.07 & 0.03 & 0.90 & 0.10 \\
 & $O{+}I$ & 0.77 & 0.93 & 0.92 & 0.08 & 1.00 & 0.00 & 0.92 & 0.00 & 0.00 & 0.95 & 0.05 \\
 & $O{+}D$ & 0.78 & 0.87 & 0.84 & 0.16 & 0.29 & 0.71 & 0.85 & 0.33 & 0.00 & 0.89 & 0.11 \\
 & $I{+}D$ & 0.80 & 0.83 & 0.89 & 0.11 & 0.93 & 0.07 & 0.83 & 0.14 & 0.04 & 0.94 & 0.05 \\
 & $O{+}I{+}D$ & 0.84 & 1.00 & 0.88 & 0.12 & 1.00 & 0.00 & 0.96 & 0.00 & 0.00 & 0.96 & 0.04 \\
\midrule
\multicolumn{13}{@{}p{0.96\textwidth}@{}}{
\footnotesize
\textit{Abbreviations:}
$O$ = information omission;
$I$ = contradiction injection;
$D$ = evidence delay;
GDR = Gap Detection Rate;
IR = Inquiry Rate;
PIR = Precision Inquiry Rate;
RIR = Redundant Inquiry Ratio;
CDR = Contradiction Detection Rate;
CIR = Contradiction Ignorance Rate;
RUR = Rational Update Rate;
PCR = Premature Closure Rate;
SMR = Stubborn Maintenance Rate;
EF = Evidence Faithfulness;
FCR = Fabricated Citation Rate.
} \\
\end{tabular}
\end{adjustbox}
\endgroup
\end{table}

We used the baseline condition as the reference and analyzed the effects of three information-flow stressors and their combinations on the five process-audit nodes. Table~ \ref{tab:multimodal_process_audit_results} shows the results on the multimodal track (llm track and agent track see Appendix~ \ref{app:detail_stressor-audit}). The delta heatmaps in Figure~ \ref{fig:stressors_audit}  show that information omission, contradiction injection, and evidence delay do not lead to a uniform degradation across all metrics. Instead, they produce node-specific changes. Overall, \textbf{\textit{Contradiction Detection}} and \textbf{\textit{Diagnosis Update}} are the most sensitive nodes, whereas \textbf{\textit{Evidence Grounding}} remains relatively stable across most conditions.

For single stressors, $O$ mainly affects the construction of the initial evidential basis. Under this condition, some models show improved information-gap recognition, but this improvement does not consistently propagate to downstream reasoning nodes. In contrast, contradiction detection and diagnosis update metrics are more likely to decline. The effect of $I$ is generally weaker, and in some cases it is associated with improved information-gap awareness or greater diagnostic openness. This suggests that an explicit contradiction does not necessarily cause global degradation and may sometimes induce a more cautious reasoning pattern. Compared with $I$, $D$ has a stronger effect on dynamic reasoning nodes, particularly those related to cross-turn evidence integration and diagnostic revision after new evidence becomes available.

Among the combined stress conditions, $O+D$ produces the most prominent impact. Under this setting, all three models show varying degrees of decline in \textbf{\textit{Contradiction Detection}} and \textbf{\textit{Diagnosis Update}}, indicating that the combination of missing information and delayed evidence substantially increases the difficulty of dynamic reasoning. By contrast, $O+I$, $I+D$, and $O+I+D$ do not exhibit a simple additive decline. In several cases, the presence of an explicit contradiction appears to increase model sensitivity to case uncertainty, partially offsetting some of the negative effects introduced by omission or delay.


\subsection{Hallucination propagation under information-flow stressors.}
The hallucination monitoring results show that information-flow stressors mainly affect \textbf{hallucination propagation} and \textbf{contradiction-related self-correction}, rather than initial hallucination generation (The detail results see  Figure~ \ref{fig:stressors_hallu} and Appendix~\ref{tag:hallucination_results}). Across the three models, NFR and UFR change only slightly under most stress conditions, suggesting that omission, contradiction, and delay do not consistently increase the immediate generation of unsupported numerical or factual claims. In contrast, HPR and HCCR show more frequent degradation, indicating that once unsupported claims are introduced, models are more likely to preserve them across turns and reuse them in subsequent diagnostic reasoning.

Among the stressors, both information omission and evidence delay weaken hallucination control, especially through declines in HPR and HCCR. The effect of explicit contradiction is more mixed: it may sometimes make the model more cautious, but it does not reliably prevent hallucination propagation. Among combined conditions, $O+D$ is the most disruptive setting. All three models show clear degradation in HPR, HCCR, and CIHSR under this condition, indicating that when evidence is both missing and delayed, models have greater difficulty suppressing exposed hallucinations and revising subsequent reasoning.


The hallucination propagation results described above identify \textit{what} goes wrong:
unsupported claims mostly spread through cross‑turn persistence and reasoning contamination.
The five‑node process audit provides complementary evidence on \textit{why} this occurs.
Across all models and stressors, the most pronounced process‑level degradations appear in
\textbf{Contradiction Detection} and \textbf{Diagnosis Update} (Section~\ref{sec:multimodal-process}).
When a model fails to recognize an injected contradiction or to adjust its belief in light of delayed evidence,
pre‑existing unsupported claims are more likely to persist and contaminate subsequent reasoning.
In other words, deficits in contradiction‑based self‑correction and belief revision constitute the primary mechanism
through which initial hallucinations evolve into anchored diagnostic errors.
By coupling trajectory‑level hallucination tracking with node‑level behavioral auditing, we provides a complete diagnostic loop that connects observed hallucination symptoms to their cognitive antecedents.


\subsection{Reliability Analysis}
\label{sec:reliability}

To assess the reliability of the automatic evaluation, we conducted an
algorithm--human agreement analysis on 440 samples randomly selected from the open-domain datasets of CCR. Each sample was independently rated by human annotators using a 5-point Likert scale by comparing the model answer with the paired reference answer. The detailed Likert-5 rubric and scoring instructions are provided in Appendix~\ref{tab:likert-5}. The inter-rater reliability among the three annotators, measured by
intraclass correlation (ICC(A,k)) \citep{mcgraw1996forming}, was 0.74 (95\% CI: 0.66--0.79),
indicating good consistency of human judgments.

For agreement analysis, the continuous algorithm scores ranging from 0 to 100 were discretized into five equal-width levels to match the human 5-point Likert ratings. Spearman's rank correlation between the raw algorithm scores and human ratings was$\rho = 0.26$ ($p < 0.001$).
Quadratic weighted Cohen's \citep{cohen1968weighted} $\kappa$ was $0.32$ (95\% bootstrap CI \citep{tibshirani1993introduction}:
$0.20$--$0.43$), indicating fair agreement between the automatic evaluation and human judgments \citep{landis1977measurement}.

\section{Related works}
Existing work has increasingly recognized that evaluating medical LLMs solely on static, outcome-based metrics is insufficient, and that process-oriented assessment under conditions of diagnostic uncertainty is essential. Long et al. \citep{long2026strongreasoningisntenough} introduced EviMed with Information Coverage Rate (ICR) to quantify evidence elicitation in interactive consultation, finding that strong diagnostic reasoning does not guarantee effective information collection, while Li et al. \citep{li2024mediq} proposed MediQ where the system refrains from making diagnostic decisions when unconfident and elicits missing details via follow-up questions. Regarding diagnostic updating under sequential evidence, Pan et al. \citep{pan2026ddxtracebenchmarkmedicaldiagnostic} introduced DDX-TRACE, demonstrating that final diagnosis scores can misrepresent workup quality as models may guess plausible diagnoses without essential evidence or update uncertainty poorly, while a measurement study \citep{wang2026measuring} documented "Convergence Regression"—models correctly identifying diagnoses at intermediate stages but abandoning them when subsequent evidence triggers pattern-matching to alternatives—creating a 30\% Access-Stability Dissociation invisible under single-shot evaluation. For evidence grounding, Ma et al. \citep{ma2026citevqabenchmarkingevidenceattribution} introduced CiteVQA with Strict Attributed Accuracy (SAA), revealing "Attribution Hallucination" where models produce correct answers while citing incorrect evidence, and Fan et al. \citep{fan2026halluhardhardmultiturnhallucination} proposed HalluHard operationalizing groundedness through inline citations across high-stakes domains including medical guidelines. Collectively, these studies have advanced process-level evaluation in specific dimensions, yet no existing framework systematically audits information gap awareness, questioning strategy, contradiction detection, diagnosis updating, and evidence fidelity within a unified architecture.

\section{Conclusion}

We introduced \textbf{MedBench v5}, a dynamic, process-oriented, and hallucination-aware benchmark for clinical multimodal AI evaluation. Unlike static medical QA benchmarks that focus mainly on final-answer correctness, MedBench v5 combines Clinical Cognitive Responsiveness and Medical Atomic Skills to evaluate both broad clinical capabilities and executable agentic skills across 18 capability areas and 63 tasks.

MedBench v5 further introduces a stress-audit-tracing protocol with three switchable information-flow stressors: information omission, contradiction injection, and evidence delay. Through a five-node process audit, the benchmark localizes failures in information gap detection, follow-up strategy, contradiction detection, diagnosis update, and evidence grounding. Our results show that stressors cause node-specific degradation, with contradiction detection and diagnosis update being especially sensitive, while final evidence grounding can remain superficially stable.

We also proposed hallucination propagation monitoring to trace unsupported claims across multi-turn trajectories. The results show that hallucination risk mainly arises from cross-turn persistence, reasoning contamination, and failed contradiction-based correction, rather than from initial fabrication alone. Overall, MedBench v5 bridges static medical QA and realistic clinical workflow evaluation, providing a unified framework for capability profiling, stress testing, process auditing, and hallucination trajectory analysis.

\section{Acknowledgment}
Supported by Shanghai Artificial Intelligence Laboratory

\bibliographystyle{unsrtnat}
\bibliography{references}  

\appendix
\section{Mutimodal tasks for Clinical Cognitive Responsiveness and Medical Atomic Skills}
\label{tab: detail_ccr_mas}
\subsection{Clinical Cognitive Responsiveness}
The databases of Clinical Cognitive Responsiveness are listed in table \ref{tab:medbench_v5_ccr}
\begin{longtable}{
  >{\small\raggedright\arraybackslash}p{3.3cm}
  >{\small\raggedright\arraybackslash}p{2.3cm}
  >{\small\raggedright\arraybackslash}p{2.3cm}
  >{\small\raggedright\arraybackslash}p{7.1cm}
}
\caption{Overview of Clinical Cognitive Responsiveness} \label{tab:medbench_v5_ccr} \\
\toprule
\textbf{Dimension} & \textbf{Dataset} & \textbf{Metrics} & \textbf{Description} \\
\midrule
\endfirsthead
\multicolumn{4}{c}{\textit{Table \thetable{} -- continued from previous page}} \\
\toprule
\textbf{Dimension} & \textbf{Dataset} & \textbf{Metrics} & \textbf{Description} \\
\midrule
\endhead
\midrule
\multicolumn{4}{r}{\textit{Continued on next page}} \\
\endfoot
\bottomrule
\endlastfoot

\multirow{10}{=}{Medical Knowledge QA}
    & MedExam & Accuracy & Covers basic, professional, and public medical subjects. All items are objective multiple-choice questions, including K-type (best single-answer) and S-type (clinical vignette best-answer) formats. \\
    & MedHC & Macro-Recall \& LLM-as-a-Judge & Health consultation dataset covering common medical examinations (internal medicine, surgery, lab tests, genetic testing) across cardiovascular, oncology, metabolic, and nutritional diseases. \\
    & MedMC & Macro-Recall \& LLM-as-a-Judge & Medication consultation dataset spanning 31 clinical departments and 320 diseases, covering disease names, drug names, indications, and treatment regimens. \\
    & MedSpeQA & Macro-Recall \& LLM-as-a-Judge & Specialty-specific QA dataset covering oncology, cardiology, respiratory, gastroenterology, and imaging. Questions involve symptoms, signs, history, medications, and family history. \\
    & MedHG & Accuracy & Triage/department recommendation dataset. Given doctor--patient dialogues about symptoms and history, the model must recommend the appropriate clinical department from a predefined set. \\
    & MedLitQA & Macro-Recall \& LLM-as-a-Judge & Medical literature comprehension and reasoning QA, evaluating the model's ability to extract knowledge and perform logical reasoning based on real clinical literature excerpts. \\
    & MedRehab & Macro-Recall \& LLM-as-a-Judge & Post-discharge rehabilitation management dataset, evaluating the model's ability to deliver personalized rehabilitation plans and guidance across rehabilitation domains. \\
    & MedRxPlan & Macro-Recall \& LLM-as-a-Judge & Precision medication education dataset in a case-based QA format, covering 10 major organ systems (respiratory, circulatory, digestive, urinary, neurological, endocrine, reproductive, musculoskeletal, immune, hematological). \\
    & MedPsychCare & Macro-Recall \& LLM-as-a-Judge & Open-domain QA dataset focusing on psychological counseling and emotional support, covering common mental health issues, crisis intervention, chronic psychological problems, and cross-scenario psychological needs. \\
    & MedPsychQA & Macro-Recall \& LLM-as-a-Judge & Psychological knowledge QA dataset designed for public mental health literacy, evaluating accurate responses across 12 psychology knowledge domains. \\
\midrule

\multirow{5}{=}{Medical Language Generation}
    & MedRecordGen & Macro-Recall \& LLM-as-a-Judge & Open-domain QA evaluating the generation of structured, standardized clinical records from doctor--patient interactions, covering outpatient records, admission notes, and discharge summaries. \\
    & MedPopular & LLM-as-a-Judge & Health science communication generation, covering five core public health domains: common disease education, chronic disease management, public health, special populations, and general health knowledge. \\
    & MedSummary & Macro-Recall \& LLM-as-a-Judge & Clinical document summarization evaluating both completeness of clinical information and conciseness, covering inpatient records, outpatient records, lab/imaging reports, surgical notes, and discharge summaries. \\
    & MedExplain & Macro-Recall \& LLM-as-a-Judge & Medical terminology explanation for patient comprehension, balancing scientific accuracy with layperson readability to bridge doctor--patient communication gaps. \\
    & MedTeach & LLM-as-a-Judge & Clinical teaching case generation from real patient records, covering five scenario types while ensuring information completeness, privacy protection, and pedagogical alignment. \\
\midrule

\multirow{10}{=}{Complex Medical Reasoning}
    & CMB-Clin-extended & Macro-Recall \& LLM-as-a-Judge & Based on complex real-world clinical records, evaluating the model's ability to apply medical knowledge for diagnosis and treatment in authentic clinical scenarios. \\
    & DDx-advanced & Accuracy & Multiple-choice questions derived from real patient records (per SCARE guidelines), covering demographics, symptoms, clinical concerns, treatment/surgical history, medications, allergies, family history, and lifestyle factors. Correct answers may include one or multiple options. \\
    & MedTreat & Macro-Recall \& LLM-as-a-Judge & Precision treatment planning for specific diseases in complex clinical scenarios, evaluating standardized and personalized therapeutic recommendations. \\
    & MedOutcome & Accuracy \& Macro-Recall \& LLM-as-a-Judge & Clinical outcome prediction (cured, improved, unchanged, deceased) based on patient information, key interventions, and disease-specific clinical guidelines. \\
    & MedAnalysis & Macro-Recall \& LLM-as-a-Judge & Personalized risk assessment using clinical scoring systems and formulas (e.g., APACHE II, CHADS\textsubscript{2}) to evaluate disease progression, treatment complications, and prognosis risks. \\
    & MedDiag & Macro-Recall \& LLM-as-a-Judge & Primary care diagnosis recommendation covering common respiratory, digestive, and community-acquired diseases, requiring diagnosis and evidence based on patient information and basic examinations. \\
    & MedDiffer & Macro-Recall \& LLM-as-a-Judge & Differential diagnosis recommendation for primary care, covering common diseases and requiring differential results with supporting evidence. \\
    & MedCare & Macro-Recall \& LLM-as-a-Judge & Appropriate treatment and management recommendations for general practitioners, covering symptomatic treatment of common diseases and long-term chronic disease management. \\
    & MedPrimary & Macro-Recall \& LLM-as-a-Judge & Clinical decision support for general practitioners in inpatient settings, requiring further diagnostic and therapeutic plans based on patient records and basic examination results. \\
    & MedPHM & Macro-Recall \& LLM-as-a-Judge & Personalized health management for chronic disease patients, covering diet/exercise planning, lifestyle intervention, acute exacerbation prevention/treatment, and treatment efficacy management for hypertension, diabetes, and COPD. \\
\midrule

\multirow{9}{=}{Medical Language Understanding}
    & SMDoc & Accuracy & Medical text structuring from real clinical documents (patient demographics, symptom descriptions, lab/imaging results), evaluating extraction of specific clinical entities such as vital signs and examination findings. \\
    & MedRxCheck & Accuracy \& Macro-Recall \& LLM-as-a-Judge & Pre-prescription intelligent review covering single-choice, multiple-choice, and open-ended questions, evaluating the model's ability to audit prescription rationality, safety, and compliance. \\
    & MedInsureCheck & Macro-Recall \& LLM-as-a-Judge & Automated audit of medical insurance claims for compliance and reasonableness based on simulated insurance data and healthcare insurance regulations. \\
    & MedInsureCalc & Accuracy & Medical insurance fee calculation and payment management across diverse scenarios, including basic settlement, complex rules (e.g., Category-B self-pay first, deductible-ceiling interaction), robustness testing, and edge cases. \\
    & MedChartQC & Macro-Recall \& LLM-as-a-Judge & Medical document quality control evaluating completeness, standardization, and logical consistency across admission records, progress notes, surgical records, discharge summaries, and outpatient records. \\
    & MedReportQC & Micro-F1 \& Macro-Recall \& LLM-as-a-Judge & CT imaging report quality control, evaluating the model's knowledge of common disease CT manifestations and diagnostic accuracy to reduce manual QC workload. \\
    & MedPathQC & Macro-Recall \& LLM-as-a-Judge & Clinical pathway quality control across the full chain of admission determination, process monitoring, and variance management, supporting standardized and homogeneous care delivery. \\
    & MedTerm & Macro-Recall \& LLM-as-a-Judge & Precision interpretation of core clinical and research medical terminology for healthcare professionals. \\
    & MedSynonym & Macro-Recall \& LLM-as-a-Judge & Multi-scenario medical synonym matching (clinical consultations, record writing, surgical communication, lab report interpretation, research writing, academic presentations, doctor--patient communication) across six domains: basic medicine, clinical diagnosis, disease pathology, therapeutic intervention, pharmacology, and laboratory medicine. \\
\midrule

\multirow{2}{=}{Healthcare Safety \& Ethics}
    & MedSafety & Accuracy & Multiple-choice questions based on healthcare quality and safety core regulations, laws, and industry standards, covering the full spectrum of clinical safety scenarios. \\
    & MedEthics & Accuracy & Single-choice questions built from classic medical ethics textbooks and domestic/international policies, covering clinical ethics, research ethics, genetics and reproductive ethics, psychiatric ethics, end-of-life care ethics, interpersonal ethics, public health ethics, traditional Chinese medicine ethics, health management ethics, organ transplantation ethics, and pandemic ethics. \\
\midrule

\multirow{3}{=}{Medical Visual Perception \& Text Extraction}
    & MedDetect & IoU \& Accuracy & Target detection in medical images (CT, MRI, etc.) guided by clinical text descriptions, evaluating the model's ability to localize and identify imaging targets. \\
    & MedClass & Accuracy & Multi-modal image classification fusing medical imaging (CT, MRI) with clinical text (chief complaints, history summaries, lab indicators) for precise image-level classification. \\
    & MedOCR & 1-N.E.D. (Normalized Edit Distance) & Named entity recognition on medical imaging reports; accurate text recognition from report images is the prerequisite for downstream content understanding. \\
\midrule

\multirow{5}{=}{Cross-modal Semantic Understanding \& Reasoning}
    & MedVQA & Macro-Recall \& LLM-as-a-Judge & Visual question answering combining imaging report text with patient demographics to provide preliminary diagnosis and further reasoning recommendations. \\
    & MedGen & Macro-Recall \& LLM-as-a-Judge & Multi-modal report generation from paired image--report data (ultrasound, X-ray, pathology, endoscopy), evaluating both content accuracy and linguistic fluency of generated clinical reports. \\
    & MedQC & Micro-F1 & Chest X-ray image quality control covering 11 QC dimensions including artifacts and improper positioning, using constrained-domain answers. \\
    & MedSeqIm & Macro-Recall \& LLM-as-a-Judge & Longitudinal imaging sequence understanding with clinical and temporal annotations, evaluating analysis of imaging changes, treatment response prediction, and multi-task temporal reasoning. \\
    & Med3DMTVQA & Accuracy & 3D multi-timepoint visual QA based on real radiology reports (CT plain scan, contrast-enhanced CT, and follow-up scans), evaluating 3D volume data + multi-sequence + temporal comparison understanding. \\
\midrule

\multirow{4}{=}{Clinical Decision Support \& Reasoning}
    & MedDiffDx & Macro-Recall \& LLM-as-a-Judge & Multi-modal clinical case differential diagnosis, evaluating the generation of probabilistic differential lists with supporting evidence and diagnostic accuracy. \\
    & MedTherapy & Macro-Recall \& LLM-as-a-Judge & Multi-modal treatment planning with annotated treatment regimens and outcomes, evaluating personalized therapy recommendations with multi-dimensional justification. \\
    & MedCourse & Macro-Recall \& LLM-as-a-Judge & Chronic disease longitudinal follow-up with multi-modal data (imaging, treatment records, complete disease course annotations), evaluating disease progression analysis and individualized modeling. \\
    & MedRealMM & LLM-as-a-Judge (case-specific rubric) & Real-world telemedicine dialogues from de-identified Chinese online consultation platforms, with multi-turn conversations and patient-uploaded medical images; models generate physician responses at key decision nodes. \\
\midrule

\multirow{4}{=}{Clinical Task Planning \& Reasoning}
    & MedDecomp & LLM-as-a-Judge & Clinical goal decomposition across five major clinical scenarios (outpatient, emergency, etc.), evaluating the model's ability to decompose abstract clinical goals into executable, logical, and comprehensive task sequences. \\
    & MedPathPlan & LLM-as-a-Judge & Multi-department complex clinical pathway planning with full-process data and multiple care pathways, evaluating the generation of compliant and personalized clinical pathways. \\
    & MedCOT & LLM-as-a-Judge & Chain-of-thought reasoning on multi-step complex clinical cases across outpatient, emergency, and chronic disease management scenarios, evaluating logical consistency and accuracy of reasoning. \\
    & MedReflect & LLM-as-a-Judge & Error-annotated clinical decision cases across four major scenarios, with initial plans, error descriptions, and expert corrections, evaluating the model's ability to identify biases/errors and propose reasonable improvements. \\
\midrule

\multirow{2}{=}{Medical Tool Invocation \& Execution}
    & MedRetAPI & LLM-as-a-Judge & Clinical information retrieval across 8 major clinical information-need scenarios for both providers and patients, evaluating query generation accuracy for information access. \\
    & MedCallAPI & LLM-as-a-Judge & External system API invocation for clinical operations across 6 scenarios with operational and parameter requirements, evaluating compliant API call generation. \\
\midrule

\multirow{2}{=}{Medical Scenario Perception \& Interaction}
    & MedIntentID & Accuracy & Multi-scenario doctor--patient dialogue intent recognition covering 6 dialogue types with full context, evaluating classification accuracy and contextual understanding. \\
    & MedRoleAdapt & LLM-as-a-Judge & Multi-role medical dialogue adaptation (patient, physician, etc.) with role-specific information and communication goals, evaluating role-appropriate and adaptive response generation. \\
\midrule

\multirow{2}{=}{Memory \& Context Retention}
    & MedLongConv & LLM-as-a-Judge & Long-term conversational tracking across three interaction types---chronic disease management, common disease management, and rehabilitation guidance---each covering multiple diseases. \\
    & MedLongQA & LLM-as-a-Judge & Long-document medical QA covering clinical records, research literature, and complex queries, evaluating the model's ability to synthesize information across lengthy documents with deep comprehension and answer accuracy. \\
\midrule

\multirow{1}{=}{Medical Multi-Agent Collaboration}
    & MedCollab & LLM-as-a-Judge & Multi-system collaborative medical scenarios covering five collaboration modes: diagnostic assistance, treatment execution, chronic disease management, emergency coordination, and rehabilitation guidance, evaluating task decomposition and system coordination. \\
\bottomrule

\end{longtable}

\subsection{Medical Atomic Skills}
The description of Medical Atomic Skills is listed in table \ref{tab:atomic_skills}

\begin{table}[H]
\centering
\caption{Overview of Agent Atomic Skill Evaluation Datasets}
\label{tab:atomic_skills}
\small
\setlength{\tabcolsep}{4pt}
\begin{tabular}{
  >{\raggedright\arraybackslash}p{2.4cm}
  >{\raggedright\arraybackslash}p{2.8cm}
  >{\raggedright\arraybackslash}p{9.5cm}
}
\toprule
\textbf{Dataset} & \textbf{Metrics} & \textbf{Description} \\
\midrule

DataAgent & Accuracy &
Comprehensively evaluates the agent's multi-dimensional query capabilities, including: (1) precise information extraction based on specific patient IDs or report IDs; (2) time-series queries that interpret natural language temporal expressions to identify the ``most recent'' test result; and (3) multi-condition composite queries involving multiple logical constraints. \\

DataAgent & MAP \& LLM-as-a-Judge &
Evaluates the model's multi-dimensional information processing and output capability within a retrieval-augmented generation (RAG) architecture. The dataset focuses on the agent's knowledge fusion ability, requiring the model to perform deduplication, denoising, segment reordering, contextual association, and conflict resolution within complex texts containing both correct information and distractor data. It also assesses the agent's query reformulation and re-ranking capabilities to ensure the retrieved knowledge best matches the user's medical question. Furthermore, the agent must explicitly and accurately cite the source document and specific passage when generating the final answer. \\

DeepResearch & SemRec \& LLM-as-a-Judge &
Contains topic-specific research tasks for research-oriented agents, covering: problem analysis and planning, requirement decomposition, multi-source information retrieval, timeliness filtering, evidence chain construction, logical reasoning, structured report generation, risk alerts, and reference alignment. Each case provides a research topic, time range, retrieval constraints, and key expected report elements. The model must plan the report structure around the topic, invoke retrieval capabilities to obtain reliable information, and integrate evidence to produce a complete research report. The dataset evaluates the model's information gathering ability, reasoning quality, report generation quality, and citation reliability. \\

SafetyAgent & Defense Success Rate &
A comprehensive medical AI safety and defense evaluation benchmark that assesses the defense capability of medical LLMs in complex real-world scenarios from two dimensions: regulatory compliance risk and technical adversarial attacks. For risk identification, the dataset covers 20 sub-category scenarios across medical ethics (11 categories: clinical diagnosis and treatment, reproductive medicine, public health, etc.) and medical safety (9 categories: medication safety, adverse event handling, information security, etc.), requiring the model to accurately identify potential medical risks in context and take appropriate mitigation measures. For system defense, the dataset incorporates over ten types of adversarial attack samples targeting medical AI systems, including prompt injection, malicious instructions, data poisoning, privilege escalation, and API interface attacks. Through comparison of normal and adversarial samples, it rigorously tests the model's system robustness and defense effectiveness. \\

\bottomrule
\end{tabular}
\end{table}

\section{Likert-5 for Human Rating}
The detail of Likert-5 for Human Rating is listed in table \ref{tab:likert-5}
\begin{table}[htbp]
\caption{Medical Relevance Rating Scale (Likert-5)}
\label{tab:likert-5}
\begin{tabularx}{\textwidth}{c c X}
\toprule
\textbf{Score} & \textbf{Level} & \textbf{Detailed Definition (Medical Relevance Context)} \\
\midrule
5 & Completely Consistent & The model answer is completely consistent with the reference answer in all key medical information (e.g., diagnosis, medication, dosage, tests, treatment principles), with no semantic differences; even if wording differs, it does not cause any ambiguity or deviation. \\
4 & Highly Consistent & The model answer shares the same core medical views as the reference answer, but has one of the following minor differences: omits non-critical supplementary information; slightly different order or level of detail without substantially affecting clinical judgment; uses synonyms or approximate terms. \\
3 & Partially Consistent & The model answer contains most of the key information of the reference answer, but simultaneously has at least one of: omits 1–2 moderately important points; contains a small amount of imprecise description; logic is roughly correct but details are vague or omitted. \\
2 & Slightly Consistent & The model answer shares only a few common points with the reference answer: covers only 1–2 of the most core points, missing a large amount of necessary information; contains clear medical factual errors; logic has obvious jumps or contradictions. \\
1 & Inconsistent & The model answer is essentially unrelated or completely wrong: contains none of the correct key information; contains serious medical errors (may mislead treatment); irrelevant to the question. \\
\bottomrule
\end{tabularx}
\end{table}

\noindent\textbf{Scoring Instructions:}
\begin{itemize}
    \item \textbf{Rating object:} Compare each model answer with its paired reference answer per question.
    \item \textbf{Blinding:} The model source should not be known during rating; only compare text content.
    \item \textbf{Handling uncertainty:} If the reference answer is incomplete, use recognized medical guidelines or clinical routine as supplementary basis. Additional correct information provided by the model is not penalized, but consistency scoring remains based on the core overlap.
    \item \textbf{Use of scores:} This rating is an ordinal categorical variable. Subsequent analyses include inter-rater consistency (e.g., Kappa coefficient) and correlation with automatic metrics (e.g., Spearman correlation coefficient).
\end{itemize}


\section{Monitoring Hallucination Propagation}
\label{subsec:hallucination_monitoring}

In addition to the five-node process audit, we monitor how unsupported claims are initiated, propagated, anchored, and modulated by contradictions under different information-flow stressors. Unlike conventional hallucination evaluation, which typically checks only whether the final answer contains unsupported content, our protocol traces hallucination behavior throughout the full multi-turn trajectory. This allows us to distinguish early fabrication, downstream reuse of hallucinated content, final-decision contamination, and hallucinations induced or suppressed by explicit contradictions.

For each scenario $(x_i,s)$ and turn $t$, the judge extracts factual claims from the model response and compares them against the information available up to that turn. Let $R_{i,t}^s$ be the set of released clinical facts, $Z_{i,t}^s$ the set of factual claims extracted from the model response, and $H_{i,t}^s$ the subset of unsupported claims:
\[
H_{i,t}^s=\{z \in Z_{i,t}^s \mid z \not\preceq R_{i,t}^s\}.
\]
Here, $z \not\preceq R_{i,t}^s$ means that the claim is not supported by the available scenario context, is contradicted by the context, or is inconsistent with the gold-standard trajectory. Based on these hallucination events, the judge produces count-based annotations that are converted into eight ratio-based metrics grouped into four dimensions: initiation, propagation, anchoring, and hallucination--contradiction interaction. When the denominator of a metric is zero, the metric is treated as not applicable for that sample and excluded from macro-averaging.

\paragraph{Initiation.}
The initiation dimension measures whether hallucinations are generated when relevant information is absent, withheld, or not yet released. We use two metrics. The \textbf{numerical fabrication ratio} (NFR) captures fabricated clinical numbers, such as invented vital signs, laboratory values, imaging measurements, or medication doses. The \textbf{unsubstantiated fact ratio} (UFR) captures unsupported non-numerical factual claims, such as invented medical history, medication use, allergies, procedures, or prior diagnoses:
\begin{equation}
\mathrm{NFR}
=
\frac{
\#\mathrm{fabricated\_numeric\_claims}
}{
\#\mathrm{total\_numeric\_claims}
},
\end{equation}
\begin{equation}
\mathrm{UFR}
=
\frac{
\#\mathrm{unsupported\_factual\_claims}
}{
\#\mathrm{total\_factual\_claims}
}.
\end{equation}

\paragraph{Propagation.}
The propagation dimension measures whether hallucinated content persists across turns or contaminates later reasoning. The \textbf{hallucination persistence ratio} (HPR) measures the proportion of initiated hallucinations that reappear in subsequent turns. The \textbf{hallucination cross-contamination ratio} (HCCR) measures whether hallucinated claims are reused as premises for later diagnostic reasoning:
\begin{equation}
\mathrm{HPR}
=
\frac{
\#\mathrm{persistent\_hallucinations}
}{
\#\mathrm{initiated\_hallucinations}
},
\end{equation}
\begin{equation}
\mathrm{HCCR}
=
\frac{
\#\mathrm{cross\_contamination\_events}
}{
\#\mathrm{cross\_contamination\_opportunities}
}.
\end{equation}

\paragraph{Anchoring.}
The anchoring dimension measures whether hallucinated content becomes fixed in the final diagnostic conclusion. The \textbf{definitive hallucination dependency ratio} (DHDR) measures the fraction of final evidence items that are hallucinated or unsupported. The \textbf{critical hallucination omission} (CHO) measures whether the model omits genuine critical evidence, especially evidence that would contradict or weaken a hallucinated reasoning path:
\begin{equation}
\mathrm{DHDR}
=
\frac{
\#\mathrm{hallucinated\_final\_evidence}
}{
\#\mathrm{final\_evidence\_items}
},
\end{equation}
\begin{equation}
\mathrm{CHO}
=
\frac{
\#\mathrm{critical\_hallucination\_omissions}
}{
\#\mathrm{critical\_evidence\_opportunities}
}.
\end{equation}

\paragraph{Hallucination--Contradiction Interaction.}
The final dimension evaluates how hallucinations interact with explicit contradictions. The \textbf{contradiction-induced hallucination suppression ratio} (CIHSR) measures whether contradictions help the model suppress or revise previously exposed hallucinations, for example by asking for clarification, acknowledging inconsistency, or withdrawing unsupported assumptions. In contrast, the \textbf{contradiction-induced hallucination generation ratio} (CIHGR) measures whether contradictions trigger new hallucinations, such as fabricated explanations introduced to reconcile inconsistent evidence:
\begin{equation}
\mathrm{CIHSR}
=
\frac{
\#\mathrm{contradiction\_suppressed\_hallucinations}
}{
\#\mathrm{hallucinations\_exposed\_by\_contradictions}
},
\end{equation}
\begin{equation}
\mathrm{CIHGR}
=
\frac{
\#\mathrm{contradiction\_induced\_new\_hallucinations}
}{
\#\mathrm{contradiction\_events}
}.
\end{equation}

Table~\ref{tab:hallucination_metrics} summarizes the eight hallucination propagation metrics. Except for CIHSR, where a higher value indicates stronger self-correction after contradiction exposure, lower values indicate fewer hallucination-related failures.

\begin{table}[t]
\centering
\caption{Hallucination propagation metrics used in the dynamic audit protocol.}
\label{tab:hallucination_metrics}
\resizebox{\linewidth}{!}{
\begin{tabular}{llll}
\toprule
Dimension & Metric & Formula & Direction \\
\midrule
Initiation 
& NFR 
& $\frac{\#\mathrm{fabricated\_numeric\_claims}}{\#\mathrm{total\_numeric\_claims}}$ 
& $\downarrow$ \\
Initiation 
& UFR 
& $\frac{\#\mathrm{unsupported\_factual\_claims}}{\#\mathrm{total\_factual\_claims}}$ 
& $\downarrow$ \\
Propagation 
& HPR 
& $\frac{\#\mathrm{persistent\_hallucinations}}{\#\mathrm{initiated\_hallucinations}}$ 
& $\downarrow$ \\
Propagation 
& HCCR 
& $\frac{\#\mathrm{cross\_contamination\_events}}{\#\mathrm{cross\_contamination\_opportunities}}$ 
& $\downarrow$ \\
Anchoring 
& DHDR 
& $\frac{\#\mathrm{hallucinated\_final\_evidence}}{\#\mathrm{final\_evidence\_items}}$ 
& $\downarrow$ \\
Anchoring 
& CHO 
& $\frac{\#\mathrm{critical\_hallucination\_omissions}}{\#\mathrm{critical\_evidence\_opportunities}}$ 
& $\downarrow$ \\
Contradiction interaction 
& CIHSR 
& $\frac{\#\mathrm{contradiction\_suppressed\_hallucinations}}{\#\mathrm{hallucinations\_exposed\_by\_contradictions}}$ 
& $\uparrow$ \\
Contradiction interaction 
& CIHGR 
& $\frac{\#\mathrm{contradiction\_induced\_new\_hallucinations}}{\#\mathrm{contradiction\_events}}$ 
& $\downarrow$ \\
\bottomrule
\end{tabular}
}
\end{table}

For each stress condition, we first compute the above metrics at the sample level and then aggregate them across all valid samples. Undefined cases caused by zero denominators are excluded from the corresponding metric average rather than being counted as zero. This prevents scenarios without hallucination opportunities, contradiction events, or numeric claims from artificially lowering the estimated hallucination risk.

\section{Additional Results}

\begin{table}[htbp]
  \centering
  \caption{Proportion of datasets where each model achieved the top score.}
  \begin{tabular}{lccc}
    \toprule
    Model & LLM & LMM & AGENT \\
    \midrule
    Claude Opus 4.7 & \textcolor{red}{\textbf{0.39}} & \textcolor{red}{\textbf{0.17}} & 0.27 \\
    Qwen3.7-Max-Preview & 0.22 & 0.00 & 0.18 \\
    Kimi-K2.6 & 0.06 & 0.08 & \textcolor{red}{\textbf{0.45}} \\
    Gemini-3.1-Pro-Preview & 0.14 & 0.00 & 0.18 \\
    GPT-5.5 & 0.08 & \textcolor{red}{\textbf{0.17}} & 0.09 \\
    Grok-4.20 Beta & 0.06 & 0.08 & 0.27 \\
    Doubao-Seed-2.0-pro & 0.06 & 0.08 & 0.00 \\
    Gemini-3.5-Flash & 0.00 & \textcolor{red}{\textbf{0.17}} & 0.00 \\
    Qwen3.7-Plus & 0.00 & \textcolor{red}{\textbf{0.17}} & 0.00 \\
    DeepSeek-V4-Pro & 0.00 & 0.00 & 0.09 \\
    GLM-5.1 & 0.00 & 0.00 & 0.09 \\
    GLM-5.1  & 0.00 & 0.08 & 0.00 \\
    MedGemma 1.5 & 0.03 & 0.00 & 0.00 \\
    \bottomrule
  \end{tabular}
\end{table}


\subsection{Detail Results of CCR}
\label{app:results_of_ccr}
\normalsize   
\setlength{\tabcolsep}{4pt}   
\begin{longtable}{p{2.5cm} *{12}{c}}
\caption{Performance of models on LLm track tasks}\\
\toprule
\textbf{Task} &
\rotatebox{90}{\textbf{DeepSeek-V4-Pro}} &
\rotatebox{90}{\textbf{Doubao-Seed-2.0-pro}} &
\rotatebox{90}{\textbf{Qwen3.7-Max-Preview}} &
\rotatebox{90}{\textbf{Kimi-K2.6}} &
\rotatebox{90}{\textbf{GLM-5.1}} &
\rotatebox{90}{\textbf{GPT-5.5}} &
\rotatebox{90}{\textbf{Claude Opus 4.7}} &
\rotatebox{90}{\textbf{Gemini-3.1-Pro-Preview}} &
\rotatebox{90}{\textbf{Grok-4.20 Beta}} &
\rotatebox{90}{\textbf{MedGemma 1.5}} &
\rotatebox{90}{\textbf{MedReason}} &
\rotatebox{90}{\textbf{HuatuoGPT}} \\
\cmidrule(lr){2-13}
& \normalsize 1.6T & \normalsize N/A & \normalsize N/A & \normalsize 1T & \normalsize 744B & \normalsize N/A & \normalsize N/A & \normalsize N/A & \normalsize N/A & \normalsize 4B & \normalsize 8B & \normalsize 72B \\
\midrule
\endfirsthead

\multicolumn{13}{c}{{\tablename\ \thetable{} -- continued from previous page}} \\
\toprule
\textbf{Task} &
\rotatebox{90}{\textbf{DeepSeek-V4-Pro}} &
\rotatebox{90}{\textbf{Doubao-Seed-2.0-pro}} &
\rotatebox{90}{\textbf{Qwen3.7-Max-Preview}} &
\rotatebox{90}{\textbf{Kimi-K2.6}} &
\rotatebox{90}{\textbf{GLM-5.1}} &
\rotatebox{90}{\textbf{GPT-5.5}} &
\rotatebox{90}{\textbf{Claude Opus 4.7}} &
\rotatebox{90}{\textbf{Gemini-3.1-Pro-Preview}} &
\rotatebox{90}{\textbf{Grok-4.20 Beta}} &
\rotatebox{90}{\textbf{MedGemma 1.5}} &
\rotatebox{90}{\textbf{MedReason}} &
\rotatebox{90}{\textbf{HuatuoGPT}} \\
\cmidrule(lr){2-13}
& \normalsize 1.6T & \normalsize N/A & \normalsize N/A & \normalsize 1T & \normalsize 744B & \normalsize N/A & \normalsize N/A & \normalsize N/A & \normalsize N/A & \normalsize 4B & \normalsize 8B & \normalsize 72B \\
\midrule
\endhead

\bottomrule
\multicolumn{13}{r}{{Continued on next page}} \\
\endfoot

\bottomrule
\endlastfoot

MedExam & 90.24 & 95.19 & 90.78 & 92.65 & 92.65 & 91.18 & 87.83 & 91.98 & 86.50 & 45.86 & 41.31 & 85.16 \\
MedHC & 63.65 & 64.94 & 66.21 & 66.05 & 65.35 & 63.48 & 68.22 & 64.81 & 62.48 & 47.87 & 36.98 & 49.51 \\
MedMC & 85.14 & 82.99 & 86.61 & 84.98 & 85.64 & 78.31 & 82.97 & 86.74 & 80.43 & 51.38 & 34.05 & 56.18 \\
MedSpeQA & 75.35 & 75.17 & 77.93 & 76.57 & 77.42 & 75.49 & 77.28 & 76.28 & 74.33 & 53.46 & 43.01 & 51.53 \\
MedHG & 62.00 & 64.00 & 57.50 & 55.00 & 67.50 & 67.50 & 54.50 & 63.50 & 73.50 & 82.50 & 47.50 & 40.50 \\
MedLitQA & 53.78 & 62.67 & 65.65 & 69.48 & 68.36 & 59.74 & 72.53 & 62.11 & 65.32 & 19.79 & 28.20 & 48.47 \\
MedRehab & 62.76 & 62.27 & 64.87 & 63.69 & 64.31 & 60.54 & 65.26 & 63.81 & 62.30 & 40.69 & 34.90 & 42.60 \\
MedRxPlan & 75.94 & 74.94 & 77.22 & 78.60 & 76.45 & 77.24 & 76.71 & 75.26 & 74.25 & 55.57 & 35.87 & 50.74 \\
MedPsychCare & 60.54 & 59.67 & 62.68 & 61.20 & 62.21 & 61.54 & 61.53 & 62.18 & 59.19 & 56.14 & 42.82 & 46.81 \\
MedPsychQA & 68.80 & 68.62 & 72.82 & 69.01 & 72.12 & 68.10 & 70.23 & 71.47 & 68.63 & 53.02 & 43.63 & 49.72 \\
MedRecordGen & 72.98 & 77.36 & 76.87 & 79.11 & 77.60 & 79.69 & 81.01 & 76.38 & 74.68 & 59.08 & 54.71 & 63.54 \\
MedPopular & 81.84 & 76.87 & 78.44 & 80.44 & 77.66 & 76.33 & 84.16 & 79.80 & 81.28 & 63.87 & 59.72 & 63.96 \\
MedSummary & 75.29 & 75.90 & 74.34 & 76.31 & 75.45 & 71.86 & 81.25 & 76.04 & 80.90 & 70.07 & 48.22 & 74.05 \\
MedExplain & 56.88 & 54.39 & 57.43 & 56.26 & 56.79 & 58.28 & 57.47 & 56.86 & 55.46 & 42.45 & 40.05 & 44.59 \\
MedTeach & 98.27 & 96.67 & 99.47 & 100.00 & 99.20 & 99.87 & 100.00 & 96.27 & 98.67 & 38.93 & 66.93 & 72.40 \\
CMB-Clin-extended & 71.01 & 71.28 & 72.22 & 72.17 & 72.33 & 72.10 & 73.08 & 71.38 & 70.27 & 55.97 & 48.84 & 61.97 \\
DDx-advanced & 35.33 & 34.67 & 45.33 & 38.00 & 34.00 & 21.33 & 29.33 & 29.33 & 37.33 & 1.33 & 1.33 & 9.33 \\
MedTreat & 59.09 & 57.47 & 57.95 & 60.95 & 60.00 & 60.96 & 61.86 & 60.59 & 57.02 & 36.22 & 32.92 & 41.54 \\
MedOutcome & 46.67 & 44.73 & 46.40 & 46.87 & 47.07 & 45.53 & 47.99 & 46.67 & 50.12 & 38.55 & 39.96 & 38.44 \\
MedAnalysis & 94.33 & 94.27 & 96.93 & 97.33 & 97.00 & 95.33 & 96.07 & 98.00 & 95.13 & 68.67 & 59.73 & 79.80 \\
MedDiag & 78.27 & 78.64 & 79.26 & 78.80 & 79.53 & 78.09 & 81.29 & 78.63 & 77.06 & 64.42 & 51.32 & 66.84 \\
MedDiffer & 45.24 & 44.40 & 43.77 & 48.34 & 44.40 & 49.42 & 47.98 & 44.18 & 44.32 & 34.40 & 28.09 & 32.66 \\
MedCare & 79.90 & 75.47 & 81.39 & 79.75 & 81.87 & 80.24 & 82.12 & 83.10 & 78.34 & 75.02 & 59.76 & 72.25 \\
MedPrimary & 78.66 & 76.53 & 76.99 & 80.96 & 79.84 & 80.96 & 84.76 & 80.67 & 74.13 & 45.63 & 39.00 & 48.55 \\
MedPHM & 42.04 & 47.17 & 49.34 & 46.60 & 48.63 & 44.35 & 47.36 & 46.61 & 43.76 & 33.29 & 16.84 & 27.82 \\
SMDoc & 67.97 & 70.47 & 71.98 & 72.03 & 67.31 & 69.21 & 71.81 & 71.95 & 72.52 & 35.23 & 16.40 & 68.03 \\
MedRxCheck & 40.51 & 39.69 & 41.07 & 38.75 & 39.44 & 39.70 & 37.12 & 39.67 & 37.71 & 6.65 & 6.09 & 26.26 \\
MedInsureCheck & 67.86 & 63.73 & 67.08 & 64.84 & 64.39 & 72.50 & 74.70 & 68.72 & 73.21 & 63.11 & 62.20 & 65.74 \\
MedInsureCalc & 75.62 & 80.38 & 76.19 & 77.05 & 75.81 & 76.29 & 76.29 & 76.29 & 78.67 & 28.67 & 30.57 & 74.57 \\
MedChartQC & 44.29 & 48.94 & 55.46 & 54.69 & 53.69 & 52.30 & 55.65 & 54.01 & 52.23 & 27.46 & 19.01 & 28.24 \\
MedReportQC & 62.94 & 66.51 & 65.40 & 65.84 & 63.33 & 63.88 & 66.16 & 66.54 & 64.51 & 38.14 & 31.03 & 51.21 \\
MedPathQC & 66.56 & 58.33 & 70.99 & 68.28 & 70.73 & 70.43 & 74.09 & 69.43 & 66.73 & 46.37 & 42.34 & 42.89 \\
MedTerm & 71.62 & 71.73 & 72.61 & 71.43 & 72.51 & 71.05 & 71.74 & 71.81 & 69.02 & 54.57 & 45.62 & 51.38 \\
MedSynonym & 70.20 & 81.27 & 82.07 & 74.80 & 77.13 & 73.73 & 77.47 & 80.87 & 70.67 & 38.93 & 32.67 & 59.00 \\
MedSafety & 47.33 & 54.67 & 62.67 & 61.33 & 46.67 & 68.67 & 58.67 & 67.33 & 47.33 & 3.33 & 0.00 & 39.33 \\
MedEthics & 42.00 & 44.00 & 50.00 & 46.67 & 42.67 & 50.00 & 53.33 & 60.67 & 49.33 & 5.33 & 0.00 & 40.67 \\
\end{longtable}

\normalsize
\setlength{\tabcolsep}{4pt}

\begin{longtable}{p{2.5cm} *{10}{c}}
\caption{Performance of multimodal medical tasks}
\label{tab:medbench_vlm} \\
\toprule
\textbf{Task} &
\rotatebox{90}{\textbf{DeepSeek-V4-Pro}} &
\rotatebox{90}{\textbf{Doubao-Seed-2.0-pro}} &
\rotatebox{90}{\textbf{Qwen3.7-Plus}} &
\rotatebox{90}{\textbf{Kimi-K2.6}} &
\rotatebox{90}{\textbf{GLM-5.1}} &
\rotatebox{90}{\textbf{GPT-5.5}} &
\rotatebox{90}{\textbf{Claude Opus 4.7}} &
\rotatebox{90}{\textbf{Gemini-3.5-Flash}} &
\rotatebox{90}{\textbf{Grok-4.20 Beta}} &
\rotatebox{90}{\textbf{MedGemma 1.5}} \\
\cmidrule(lr){2-11}
& \normalsize 1600B & \normalsize N/A & \normalsize 35B & \normalsize 1000B & \normalsize 744B & \normalsize N/A & \normalsize N/A & \normalsize N/A & \normalsize N/A & \normalsize 4B \\
\midrule
\endfirsthead

\multicolumn{11}{c}{{\tablename\ \thetable{} -- continued from previous page}} \\
\toprule
\textbf{Task} &
\rotatebox{90}{\textbf{DeepSeek-V4-Pro}} &
\rotatebox{90}{\textbf{Doubao-Seed-2.0-pro}} &
\rotatebox{90}{\textbf{Qwen3.7-Plus}} &
\rotatebox{90}{\textbf{Kimi-K2.6}} &
\rotatebox{90}{\textbf{GLM-5.1}} &
\rotatebox{90}{\textbf{GPT-5.5}} &
\rotatebox{90}{\textbf{Claude Opus 4.7}} &
\rotatebox{90}{\textbf{Gemini-3.5-Flash}} &
\rotatebox{90}{\textbf{Grok-4.20 Beta}} &
\rotatebox{90}{\textbf{MedGemma 1.5}} \\
\cmidrule(lr){2-11}
& \normalsize 1600B & \normalsize N/A & \normalsize 35B & \normalsize 1000B & \normalsize 744B & \normalsize N/A & \normalsize N/A & \normalsize N/A & \normalsize N/A & \normalsize 4B \\
\midrule
\endhead

\bottomrule
\multicolumn{11}{r}{{Continued on next page}} \\
\endfoot

\bottomrule
\endlastfoot

MedDetect  & 8.74  & 27.63 & 12.89 & 8.88  & 2.13  & 10.64 & 4.59  & 27.87 & 4.98  & 11.38 \\
MedClass   & 57.86 & 65.71 & 68.57 & 67.86 & 55.00 & 65.71 & 72.86 & 69.29 & 9.29  & 3.57  \\
MedOCR     & 1.00  & 65.33 & 64.46 & 60.65 & 0.97  & 68.37 & 64.95 & 57.95 & 59.79 & 20.77 \\
MedVQA     & 32.90 & 47.11 & 44.57 & 32.15 & 45.38 & 32.93 & 38.74 & 40.70 & 26.42 & 23.43 \\
MedGen     & 35.20 & 41.25 & 41.30 & 40.44 & 37.12 & 42.16 & 42.85 & 43.08 & 37.48 & 23.16 \\
MedQC      & 11.85 & 16.34 & 13.95 & 15.12 & 13.25 & 14.35 & 12.58 & 7.35  & 17.77 & 0.86  \\
MedSeqIm   & 36.12 & 42.14 & 60.66 & 41.99 & 40.99 & 42.93 & 42.80 & 42.36 & 39.81 & 25.52 \\
MedDiffDx  & 64.08 & 65.97 & 41.99 & 48.79 & 65.80 & 63.99 & 67.61 & 63.05 & 60.02 & 46.90 \\
MedTherapy & 57.54 & 56.46 & 42.88 & 65.75 & 59.99 & 59.24 & 62.92 & 58.52 & 56.73 & 46.45 \\
MedCourse  & 69.18 & 67.09 & 67.29 & 69.25 & 73.22 & 71.07 & 72.35 & 68.72 & 68.50 & 57.39 \\
MedRealMM  & 81.93 & 82.74 & 60.59 & 68.19 & 84.97 & 86.22 & 53.15 & 84.20 & 5.12  & 46.15 \\
Med3DMTVQA & 37.74 & 42.02 & 71.67 & 37.74 & 29.33 & 43.22 & 40.31 & 45.11 & 35.16 & 31.05 \\

\end{longtable}


\begin{longtable}{p{2.5cm} *{12}{c}}
\caption{Performance of models on agent track tasks}\\
\toprule
\textbf{Task} &
\rotatebox{90}{\textbf{DeepSeek-V4-Pro}} &
\rotatebox{90}{\textbf{Doubao-Seed-2.0-pro}} &
\rotatebox{90}{\textbf{Qwen3.7-Max-Preview}} &
\rotatebox{90}{\textbf{Kimi-K2.6}} &
\rotatebox{90}{\textbf{GLM-5.1}} &
\rotatebox{90}{\textbf{GPT-5.5}} &
\rotatebox{90}{\textbf{Claude Opus 4.7}} &
\rotatebox{90}{\textbf{Gemini-3.1-Pro-Preview}} &
\rotatebox{90}{\textbf{Grok-4.20 Beta}} &
\rotatebox{90}{\textbf{MedGemma 1.5}} &
\rotatebox{90}{\textbf{MedReason}} &
\rotatebox{90}{\textbf{HuatuoGPT}} \\
\cmidrule(lr){2-13}
& \normalsize 1.6T & \normalsize N/A & \normalsize N/A & \normalsize 1T & \normalsize 744B & \normalsize N/A & \normalsize N/A & \normalsize N/A & \normalsize N/A & \normalsize 4B & \normalsize 8B & \normalsize 72B \\
\midrule
\endfirsthead

\multicolumn{13}{c}{{\tablename\ \thetable{} -- continued from previous page}} \\
\toprule
\textbf{Task} &
\rotatebox{90}{\textbf{DeepSeek-V4-Pro}} &
\rotatebox{90}{\textbf{Doubao-Seed-2.0-pro}} &
\rotatebox{90}{\textbf{Qwen3.7-Max-Preview}} &
\rotatebox{90}{\textbf{Kimi-K2.6}} &
\rotatebox{90}{\textbf{GLM-5.1}} &
\rotatebox{90}{\textbf{GPT-5.5}} &
\rotatebox{90}{\textbf{Claude Opus 4.7}} &
\rotatebox{90}{\textbf{Gemini-3.1-Pro-Preview}} &
\rotatebox{90}{\textbf{Grok-4.20 Beta}} &
\rotatebox{90}{\textbf{MedGemma 1.5}} &
\rotatebox{90}{\textbf{MedReason}} &
\rotatebox{90}{\textbf{HuatuoGPT}} \\
\cmidrule(lr){2-13}
& \normalsize 1.6T & \normalsize N/A & \normalsize N/A & \normalsize 1T & \normalsize 744B & \normalsize N/A & \normalsize N/A & \normalsize N/A & \normalsize N/A & \normalsize 4B & \normalsize 8B & \normalsize 72B \\
\midrule
\endhead

\bottomrule
\multicolumn{13}{r}{{Continued on next page}} \\
\endfoot

\bottomrule
\endlastfoot

MedDecomp & 95.47 & 96.53 & 98.13 & 99.33 & 97.47 & 96.13 & 99.73 & 93.20 & 97.33 & 79.20 & 69.33 & 72.27 \\
MedPathPlan & 94.67 & 92.67 & 94.27 & 92.40 & 85.07 & 95.33 & 93.20 & 90.40 & 89.47 & 82.00 & 66.00 & 77.47 \\
MedCOT & 99.80 & 99.80 & 98.80 & 100.00 & 99.90 & 99.90 & 99.90 & 95.50 & 100.00 & 93.00 & 76.30 & 80.30 \\
MedReflect & 96.20 & 92.50 & 96.60 & 99.70 & 96.30 & 99.00 & 99.30 & 94.60 & 96.60 & 77.90 & 66.10 & 73.20 \\
MedRetAPI & 98.68 & 95.13 & 99.21 & 99.87 & 97.50 & 95.00 & 99.47 & 96.97 & 89.74 & 81.71 & 77.24 & 79.61 \\
MedCallAPI & 82.52 & 95.23 & 89.93 & 96.29 & 91.26 & 89.67 & 94.44 & 92.45 & 91.79 & 85.43 & 77.35 & 88.74 \\
MedIntentID & 75.17 & 81.21 & 91.95 & 83.89 & 91.28 & 90.60 & 86.58 & 91.95 & 86.58 & 0.00 & 0.00 & 69.80 \\
MedRoleAdapt & 98.80 & 89.50 & 98.70 & 99.60 & 97.20 & 97.20 & 98.40 & 98.10 & 97.40 & 75.00 & 75.00 & 79.80 \\
MedLongConv & 98.80 & 90.50 & 98.90 & 99.00 & 98.10 & 96.70 & 99.20 & 99.10 & 97.70 & 73.50 & 57.50 & 79.90 \\
MedLongQA & 84.27 & 83.87 & 86.67 & 92.00 & 89.73 & 86.53 & 93.07 & 86.40 & 94.00 & 74.67 & 72.53 & 80.40 \\
MedCollab & 100.00 & 99.87 & 100.00 & 99.33 & 100.00 & 99.87 & 100.00 & 100.00 & 100.00 & 79.87 & 67.73 & 77.07 \\
\end{longtable}

\subsection{Detail results of stressor-audit}
\label{app:detail_stressor-audit}

\begin{table}[t]
\centering
\begingroup
\renewcommand{\arraystretch}{0.78}
\caption{Process-audit evaluation results on the \textbf{LLM track} under different information-flow stress conditions. Metrics are grouped by the five audit nodes. Upward arrows indicate higher-is-better metrics, while downward arrows indicate lower-is-better metrics.}
\label{tab:llm_process_audit_results}
\begin{adjustbox}{max width=\textwidth,max totalheight=0.72\textheight}
\begin{tabular}{@{}cc*{11}{c}@{}}
\toprule
\multirow{2}{*}{\textbf{Model}}
& \multirow{2}{*}{\textbf{Stress}}
& \multicolumn{2}{c}{\textbf{Inform. Gap}}
& \multicolumn{2}{c}{\textbf{Follow-up}}
& \multicolumn{2}{c}{\textbf{Contradiction}}
& \multicolumn{3}{c}{\textbf{Diagnosis Update}}
& \multicolumn{2}{c}{\textbf{Evid. Grd.}}
\\
\cmidrule(lr){3-4}
\cmidrule(lr){5-6}
\cmidrule(lr){7-8}
\cmidrule(lr){9-11}
\cmidrule(lr){12-13}
&
& \textbf{GDR}$\uparrow$
& \textbf{IR}$\uparrow$
& \textbf{PIR}$\uparrow$
& \textbf{RIR}$\downarrow$
& \textbf{CDR}$\uparrow$
& \textbf{CIR}$\downarrow$
& \textbf{RUR}$\uparrow$
& \textbf{PCR}$\downarrow$
& \textbf{SMR}$\downarrow$
& \textbf{EF}$\uparrow$
& \textbf{FCR}$\downarrow$ \\
\midrule
\multirow{8}{*}{\begin{tabular}[c]{@{}c@{}}claude\\opus-4.7\end{tabular}}
& Baseline & 0.99 & 1.00 & 0.90 & 0.10 & 0.00 & 1.00 & 0.97 & 0.00 & 0.00 & 0.96 & 0.04 \\
& $O$ & 0.98 & 1.00 & 0.86 & 0.14 & 1.00 & 0.00 & 0.99 & 0.02 & 0.00 & 0.94 & 0.06 \\
& $I$ & 0.94 & 1.00 & 0.87 & 0.13 & 1.00 & 0.00 & 0.96 & 0.00 & 0.02 & 0.94 & 0.06 \\
& $D$ & 0.96 & 1.00 & 0.85 & 0.15 & 1.00 & 0.00 & 0.98 & 0.00 & 0.02 & 0.94 & 0.06 \\
& $O{+}I$ & 0.95 & 1.00 & 0.83 & 0.17 & 0.99 & 0.01 & 0.98 & 0.00 & 0.00 & 0.96 & 0.04 \\
& $O{+}D$ & 0.94 & 1.00 & 0.83 & 0.17 & 1.00 & 0.00 & 0.96 & 0.04 & 0.00 & 0.94 & 0.06 \\
& $I{+}D$ & 0.92 & 1.00 & 0.86 & 0.14 & 1.00 & 0.00 & 0.99 & 0.00 & 0.00 & 0.95 & 0.05 \\
& $O{+}I{+}D$ & 0.95 & 1.00 & 0.86 & 0.14 & 0.97 & 0.03 & 0.98 & 0.00 & 0.01 & 0.96 & 0.04 \\
\midrule
\multirow{8}{*}{\begin{tabular}[c]{@{}c@{}}qwen3.7-max\\2026-05-20\end{tabular}}
& Baseline & 0.96 & 1.00 & 0.94 & 0.06 & 1.00 & 0.00 & 0.99 & 0.02 & 0.07 & 0.95 & 0.05 \\
& $O$ & 0.95 & 0.98 & 0.85 & 0.15 & 1.00 & 0.00 & 0.99 & 0.06 & 0.00 & 0.93 & 0.07 \\
& $I$ & 0.93 & 1.00 & 0.89 & 0.11 & 0.98 & 0.02 & 0.96 & 0.02 & 0.03 & 0.94 & 0.06 \\
& $D$ & 0.95 & 1.00 & 0.84 & 0.16 & 1.00 & 0.00 & 0.94 & 0.00 & 0.03 & 0.94 & 0.06 \\
& $O{+}I$ & 0.95 & 1.00 & 0.85 & 0.15 & 1.00 & 0.00 & 0.96 & 0.00 & 0.02 & 0.95 & 0.05 \\
& $O{+}D$ & 0.92 & 1.00 & 0.80 & 0.20 & 1.00 & 0.00 & 0.94 & 0.06 & 0.05 & 0.94 & 0.06 \\
& $I{+}D$ & 0.93 & 1.00 & 0.85 & 0.15 & 1.00 & 0.00 & 0.99 & 0.00 & 0.01 & 0.94 & 0.06 \\
& $O{+}I{+}D$ & 0.91 & 1.00 & 0.83 & 0.17 & 0.90 & 0.10 & 0.95 & 0.00 & 0.03 & 0.95 & 0.05 \\
\midrule
\multirow{8}{*}{\begin{tabular}[c]{@{}c@{}}gemini-3.1\\pro-preview\end{tabular}}
& Baseline & 0.99 & 1.00 & 0.94 & 0.06 & 1.00 & 0.00 & 0.94 & 0.00 & 0.09 & 0.96 & 0.04 \\
& $O$ & 0.98 & 1.00 & 0.87 & 0.13 & 1.00 & 0.00 & 0.99 & 0.02 & 0.00 & 0.94 & 0.06 \\
& $I$ & 0.96 & 1.00 & 0.90 & 0.10 & 0.98 & 0.02 & 0.95 & 0.02 & 0.02 & 0.95 & 0.05 \\
& $D$ & 0.95 & 1.00 & 0.86 & 0.14 & 1.00 & 0.00 & 0.98 & 0.00 & 0.01 & 0.94 & 0.06 \\
& $O{+}I$ & 0.94 & 1.00 & 0.85 & 0.15 & 1.00 & 0.00 & 0.97 & 0.00 & 0.02 & 0.95 & 0.05 \\
& $O{+}D$ & 0.94 & 1.00 & 0.81 & 0.19 & 1.00 & 0.00 & 0.95 & 0.04 & 0.04 & 0.91 & 0.09 \\
& $I{+}D$ & 0.90 & 1.00 & 0.85 & 0.15 & 0.99 & 0.01 & 0.96 & 0.00 & 0.01 & 0.94 & 0.06 \\
& $O{+}I{+}D$ & 0.91 & 1.00 & 0.85 & 0.15 & 0.90 & 0.10 & 0.94 & 0.00 & 0.02 & 0.94 & 0.06 \\
\midrule
\multicolumn{13}{@{}p{0.96\textwidth}@{}}{
\footnotesize
\textit{Abbreviations:}
$O$ = information omission;
$I$ = contradiction injection;
$D$ = evidence delay;
GDR = Gap Detection Rate;
IR = Inquiry Rate;
PIR = Precision Inquiry Rate;
RIR = Redundant Inquiry Ratio;
CDR = Contradiction Detection Rate;
CIR = Contradiction Ignorance Rate;
RUR = Rational Update Rate;
PCR = Premature Closure Rate;
SMR = Stubborn Maintenance Rate;
EF = Evidence Faithfulness;
FCR = Fabricated Citation Rate.
} \\
\end{tabular}
\end{adjustbox}
\endgroup
\end{table}

\begin{table}[t]
\centering
\begingroup
\renewcommand{\arraystretch}{0.78}
\caption{Process-audit evaluation results on the \textbf{agent track} under different information-flow stress conditions. Metrics are grouped by the five audit nodes. Upward arrows indicate higher-is-better metrics, while downward arrows indicate lower-is-better metrics.}
\label{tab:agent_process_audit_results}
\begin{adjustbox}{max width=\textwidth,max totalheight=0.72\textheight}
\begin{tabular}{@{}cc*{11}{c}@{}}
\toprule
\multirow{2}{*}{\textbf{Model}}
& \multirow{2}{*}{\textbf{Stress}}
& \multicolumn{2}{c}{\textbf{Inform. Gap}}
& \multicolumn{2}{c}{\textbf{Follow-up}}
& \multicolumn{2}{c}{\textbf{Contradiction}}
& \multicolumn{3}{c}{\textbf{Diagnosis Update}}
& \multicolumn{2}{c}{\textbf{Evid. Grd.}}
\\
\cmidrule(lr){3-4}
\cmidrule(lr){5-6}
\cmidrule(lr){7-8}
\cmidrule(lr){9-11}
\cmidrule(lr){12-13}
&
& \textbf{GDR}$\uparrow$
& \textbf{IR}$\uparrow$
& \textbf{PIR}$\uparrow$
& \textbf{RIR}$\downarrow$
& \textbf{CDR}$\uparrow$
& \textbf{CIR}$\downarrow$
& \textbf{RUR}$\uparrow$
& \textbf{PCR}$\downarrow$
& \textbf{SMR}$\downarrow$
& \textbf{EF}$\uparrow$
& \textbf{FCR}$\downarrow$ \\
\midrule
\multirow{8}{*}{\begin{tabular}[c]{@{}c@{}}kimi-k2.6\end{tabular}}
 & Baseline & 0.94 & 1.00 & 0.90 & 0.10 & 1.00 & 0.00 & 0.91 & 0.14 & 0.33 & 0.87 & 0.11 \\
 & $O$ & 0.86 & 0.88 & 0.85 & 0.15 & 0.50 & 0.50 & 0.87 & 0.20 & 0.10 & 0.86 & 0.14 \\
 & $I$ & 0.81 & 1.00 & 0.88 & 0.12 & 0.96 & 0.04 & 0.94 & 0.00 & 0.00 & 0.89 & 0.11 \\
 & $D$ & 0.91 & 1.00 & 0.91 & 0.09 & 1.00 & 0.00 & 0.94 & 0.12 & 0.02 & 0.89 & 0.11 \\
 & $O{+}I$ & 0.86 & 0.92 & 0.84 & 0.16 & 0.90 & 0.10 & 0.90 & 0.04 & 0.04 & 0.91 & 0.09 \\
 & $O{+}D$ & 0.92 & 0.96 & 0.80 & 0.20 & 1.00 & 0.00 & 0.95 & 0.16 & 0.04 & 0.88 & 0.11 \\
 & $I{+}D$ & 0.80 & 1.00 & 0.84 & 0.16 & 0.88 & 0.12 & 0.88 & 0.04 & 0.10 & 0.88 & 0.10 \\
 & $O{+}I{+}D$ & 0.90 & 0.96 & 0.85 & 0.15 & 0.99 & 0.01 & 0.95 & 0.04 & 0.00 & 0.92 & 0.08 \\
\midrule
\multirow{8}{*}{\begin{tabular}[c]{@{}c@{}}claude\\opus-4.7\end{tabular}}
 & Baseline & 0.96 & 1.00 & 0.88 & 0.12 & 1.00 & 0.00 & 0.95 & 0.00 & 0.00 & 0.89 & 0.11 \\
 & $O$ & 0.95 & 0.92 & 0.81 & 0.19 & 1.00 & 0.00 & 0.98 & 0.28 & 0.00 & 0.88 & 0.12 \\
 & $I$ & 0.80 & 1.00 & 0.91 & 0.09 & 0.96 & 0.04 & 0.96 & 0.00 & 0.04 & 0.92 & 0.08 \\
 & $D$ & 0.91 & 1.00 & 0.83 & 0.17 & 0.86 & 0.14 & 0.90 & 0.12 & 0.07 & 0.89 & 0.10 \\
 & $O{+}I$ & 0.90 & 0.96 & 0.81 & 0.19 & 0.94 & 0.06 & 0.95 & 0.04 & 0.04 & 0.91 & 0.09 \\
 & $O{+}D$ & 0.99 & 1.00 & 0.78 & 0.22 & 0.88 & 0.12 & 0.93 & 0.09 & 0.07 & 0.88 & 0.12 \\
 & $I{+}D$ & 0.94 & 1.00 & 0.79 & 0.21 & 0.96 & 0.04 & 0.94 & 0.00 & 0.00 & 0.91 & 0.08 \\
 & $O{+}I{+}D$ & 0.96 & 1.00 & 0.80 & 0.20 & 0.97 & 0.03 & 1.00 & 0.00 & 0.00 & 0.92 & 0.08 \\
\midrule
\multirow{8}{*}{\begin{tabular}[c]{@{}c@{}}grok-4.20\\beta-0309-reasoning\end{tabular}}
 & Baseline & 0.87 & 1.00 & 0.83 & 0.17 & 1.00 & 0.00 & 0.82 & 0.25 & 0.25 & 0.86 & 0.13 \\
 & $O$ & 0.81 & 0.83 & 0.78 & 0.22 & 1.00 & 0.00 & 0.92 & 0.42 & 0.16 & 0.86 & 0.14 \\
 & $I$ & 0.82 & 1.00 & 0.81 & 0.19 & 0.94 & 0.06 & 0.90 & 0.00 & 0.04 & 0.88 & 0.12 \\
 & $D$ & 0.82 & 0.96 & 0.80 & 0.20 & 0.83 & 0.17 & 0.86 & 0.24 & 0.09 & 0.84 & 0.16 \\
 & $O{+}I$ & 0.85 & 0.88 & 0.77 & 0.23 & 0.92 & 0.08 & 0.91 & 0.04 & 0.02 & 0.91 & 0.08 \\
 & $O{+}D$ & 0.90 & 0.92 & 0.76 & 0.24 & 0.83 & 0.17 & 0.89 & 0.16 & 0.10 & 0.84 & 0.13 \\
 & $I{+}D$ & 0.79 & 0.95 & 0.76 & 0.24 & 0.88 & 0.12 & 0.90 & 0.09 & 0.08 & 0.91 & 0.09 \\
 & $O{+}I{+}D$ & 0.85 & 0.92 & 0.76 & 0.24 & 0.90 & 0.10 & 0.87 & 0.12 & 0.05 & 0.88 & 0.10 \\
\midrule
\multicolumn{13}{@{}p{0.96\textwidth}@{}}{
\footnotesize
\textit{Abbreviations:}
$O$ = information omission;
$I$ = contradiction injection;
$D$ = evidence delay;
GDR = Gap Detection Rate;
IR = Inquiry Rate;
PIR = Precision Inquiry Rate;
RIR = Redundant Inquiry Ratio;
CDR = Contradiction Detection Rate;
CIR = Contradiction Ignorance Rate;
RUR = Rational Update Rate;
PCR = Premature Closure Rate;
SMR = Stubborn Maintenance Rate;
EF = Evidence Faithfulness;
FCR = Fabricated Citation Rate.
} \\
\end{tabular}
\end{adjustbox}
\endgroup
\end{table}

\subsection{ Detail Results of Hallucination Propagation Monitoring}
\label{tag:hallucination_results}

\begin{figure}[htbp]  
    \centering        
    \includegraphics[width=1.0\textwidth]{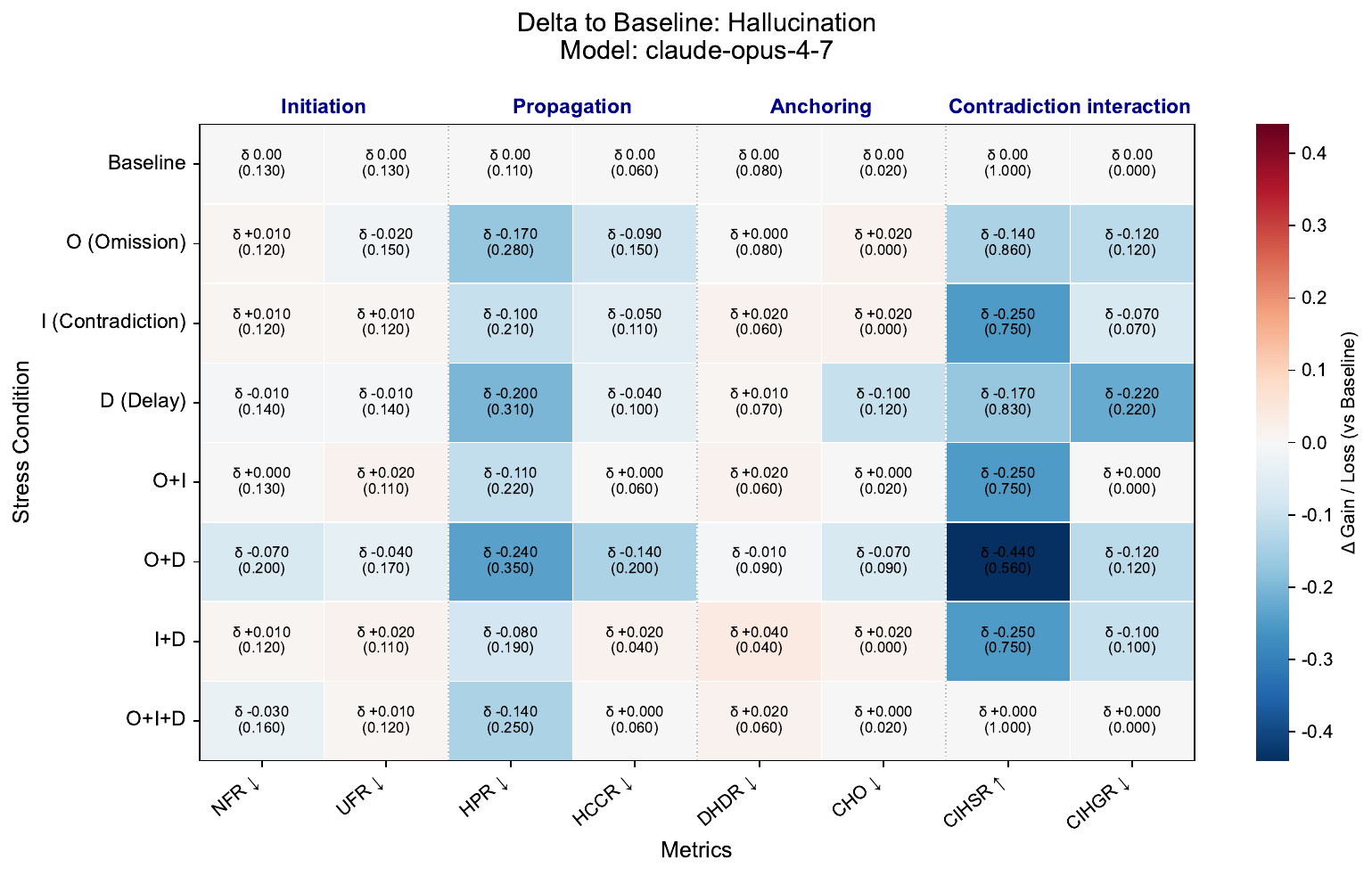}
    \caption{\textbf{Delta-to-Baseline heatmaps of Hallucination propagation metrics under stress conditions.} 
Each heatmap shows the performance change ($\Delta$) of eight stress conditions (rows) relative to the baseline (first row).
For metrics where higher is better ($\uparrow$), $\Delta = \text{metric}_{\text{stress}} - \text{metric}_{\text{baseline}}$; for metrics where lower is better ($\downarrow$), $\Delta = \text{metric}_{\text{baseline}} - \text{metric}_{\text{stress}}$.
Positive $\Delta$ (blue) indicates gain, negative $\Delta$ (red) indicates loss; baseline $\Delta$ is always zero.}  
    \label{fig:stressors_hallu} 
\end{figure}

\begin{table}[t]
\centering
\begingroup
\renewcommand{\arraystretch}{0.78}
\caption{Hallucination propagation evaluation results on the \textbf{llm track} under different information-flow stress conditions. Metrics are grouped by the four hallucination dimensions.}
\label{tab:llm_hallucination_results}
\begin{adjustbox}{max width=\textwidth,max totalheight=0.72\textheight}
\begin{tabular}{@{}cc*{8}{c}@{}}
\toprule
\multirow{2}{*}{\textbf{Model}}
& \multirow{2}{*}{\textbf{Stress}}
& \multicolumn{2}{c}{\textbf{Initiation}}
& \multicolumn{2}{c}{\textbf{Propagation}}
& \multicolumn{2}{c}{\textbf{Anchoring}}
& \multicolumn{2}{c}{\textbf{Contra. Inter.}}
\\
\cmidrule(lr){3-4}
\cmidrule(lr){5-6}
\cmidrule(lr){7-8}
\cmidrule(lr){9-10}
&
& \textbf{NFR}$\downarrow$
& \textbf{UFR}$\downarrow$
& \textbf{HPR}$\downarrow$
& \textbf{HCCR}$\downarrow$
& \textbf{DHDR}$\downarrow$
& \textbf{CHO}$\downarrow$
& \textbf{CIHSR}$\uparrow$
& \textbf{CIHGR}$\downarrow$ \\
\midrule
\multirow{8}{*}{\begin{tabular}[c]{@{}c@{}}claude\\opus-4.7\end{tabular}}
 & Baseline & 0.10 & 0.09 & 0.10 & 0.07 & 0.04 & 0.04 & 1.00 & 0.00 \\
 & $O$ & 0.10 & 0.09 & 0.17 & 0.07 & 0.05 & 0.02 & 1.00 & 0.00 \\
 & $I$ & 0.09 & 0.09 & 0.08 & 0.03 & 0.05 & 0.01 & 1.00 & 0.00 \\
 & $D$ & 0.10 & 0.09 & 0.17 & 0.07 & 0.05 & 0.03 & 0.50 & 0.00 \\
 & $O{+}I$ & 0.06 & 0.07 & 0.10 & 0.03 & 0.04 & 0.01 & 1.00 & 0.01 \\
 & $O{+}D$ & 0.11 & 0.11 & 0.20 & 0.08 & 0.06 & 0.03 & 1.00 & 0.00 \\
 & $I{+}D$ & 0.12 & 0.09 & 0.11 & 0.01 & 0.04 & 0.02 & 1.00 & 0.00 \\
 & $O{+}I{+}D$ & 0.09 & 0.08 & 0.13 & 0.02 & 0.03 & 0.04 & 1.00 & 0.01 \\
\midrule
\multirow{8}{*}{\begin{tabular}[c]{@{}c@{}}qwen3.7-max\\2026-05-20\end{tabular}}
 & Baseline & 0.09 & 0.10 & 0.19 & 0.12 & 0.05 & 0.04 & 1.00 & 0.00 \\
 & $O$ & 0.08 & 0.11 & 0.23 & 0.06 & 0.06 & 0.03 & 1.00 & 0.00 \\
 & $I$ & 0.10 & 0.11 & 0.16 & 0.04 & 0.05 & 0.03 & 0.75 & 0.04 \\
 & $D$ & 0.08 & 0.10 & 0.24 & 0.09 & 0.05 & 0.05 & 0.75 & 0.00 \\
 & $O{+}I$ & 0.05 & 0.10 & 0.13 & 0.03 & 0.05 & 0.04 & 0.67 & 0.01 \\
 & $O{+}D$ & 0.06 & 0.11 & 0.30 & 0.13 & 0.05 & 0.04 & 0.78 & 0.07 \\
 & $I{+}D$ & 0.08 & 0.09 & 0.14 & 0.04 & 0.05 & 0.02 & 0.93 & 0.05 \\
 & $O{+}I{+}D$ & 0.06 & 0.10 & 0.12 & 0.05 & 0.04 & 0.05 & 0.91 & 0.02 \\
\midrule
\multirow{8}{*}{\begin{tabular}[c]{@{}c@{}}gemini-3.1\\pro-preview\end{tabular}}
 & Baseline & 0.07 & 0.10 & 0.12 & 0.08 & 0.04 & 0.03 & 1.00 & 0.00 \\
 & $O$ & 0.09 & 0.10 & 0.21 & 0.12 & 0.05 & 0.03 & 0.50 & 0.14 \\
 & $I$ & 0.09 & 0.09 & 0.13 & 0.08 & 0.04 & 0.02 & 0.50 & 0.04 \\
 & $D$ & 0.10 & 0.11 & 0.18 & 0.09 & 0.06 & 0.03 & 0.33 & 0.22 \\
 & $O{+}I$ & 0.08 & 0.09 & 0.16 & 0.07 & 0.05 & 0.04 & 0.67 & 0.01 \\
 & $O{+}D$ & 0.10 & 0.12 & 0.23 & 0.17 & 0.08 & 0.06 & 0.57 & 0.21 \\
 & $I{+}D$ & 0.10 & 0.10 & 0.09 & 0.05 & 0.04 & 0.04 & 0.75 & 0.04 \\
 & $O{+}I{+}D$ & 0.07 & 0.10 & 0.08 & 0.05 & 0.04 & 0.08 & 0.92 & 0.06 \\
\midrule
\multicolumn{10}{@{}p{0.96\textwidth}@{}}{
\footnotesize
\textit{Abbreviations:}
$O$ = information omission;
$I$ = contradiction injection;
$D$ = evidence delay;
NFR = Numerical Fabrication Ratio;
UFR = Unsubstantiated Fact Ratio;
HPR = Hallucination Persistence Ratio;
HCCR = Hallucination Cross-Contamination Ratio;
DHDR = Definitive Hallucination Dependency Ratio;
CHO = Critical Hallucination Omission;
CIHSR = Contradiction-Induced Hallucination Suppression Ratio;
CIHGR = Contradiction-Induced Hallucination Generation Ratio.
} \\
\end{tabular}
\end{adjustbox}
\endgroup
\end{table}

\begin{table}[t]
\centering
\begingroup
\renewcommand{\arraystretch}{0.78}
\caption{Hallucination propagation evaluation results on the \textbf{multimodal track} under different information-flow stress conditions. Metrics are grouped by the four hallucination dimensions.}
\label{tab:mutimodal_hallucination_results}
\begin{adjustbox}{max width=\textwidth,max totalheight=0.72\textheight}
\begin{tabular}{@{}cc*{8}{c}@{}}
\toprule
\multirow{2}{*}{\textbf{Model}}
& \multirow{2}{*}{\textbf{Stress}}
& \multicolumn{2}{c}{\textbf{Initiation}}
& \multicolumn{2}{c}{\textbf{Propagation}}
& \multicolumn{2}{c}{\textbf{Anchoring}}
& \multicolumn{2}{c}{\textbf{Contra. Inter.}}
\\
\cmidrule(lr){3-4}
\cmidrule(lr){5-6}
\cmidrule(lr){7-8}
\cmidrule(lr){9-10}
&
& \textbf{NFR}$\downarrow$
& \textbf{UFR}$\downarrow$
& \textbf{HPR}$\downarrow$
& \textbf{HCCR}$\downarrow$
& \textbf{DHDR}$\downarrow$
& \textbf{CHO}$\downarrow$
& \textbf{CIHSR}$\uparrow$
& \textbf{CIHGR}$\downarrow$ \\
\midrule
\multirow{8}{*}{\begin{tabular}[c]{@{}c@{}}claude\\opus-4.7\end{tabular}}
  & Baseline & 0.13 & 0.13 & 0.11 & 0.06 & 0.08 & 0.02 & 1.00 & 0.00 \\
 & $O$ & 0.12 & 0.15 & 0.28 & 0.15 & 0.08 & 0.00 & 0.86 & 0.12 \\
 & $I$ & 0.12 & 0.12 & 0.21 & 0.11 & 0.06 & 0.00 & 0.75 & 0.07 \\
 & $D$ & 0.14 & 0.14 & 0.31 & 0.10 & 0.07 & 0.12 & 0.83 & 0.22 \\
 & $O{+}I$ & 0.13 & 0.11 & 0.22 & 0.06 & 0.06 & 0.02 & 0.75 & 0.00 \\
 & $O{+}D$ & 0.20 & 0.17 & 0.35 & 0.20 & 0.09 & 0.09 & 0.56 & 0.12 \\
 & $I{+}D$ & 0.12 & 0.11 & 0.19 & 0.04 & 0.04 & 0.00 & 0.75 & 0.10 \\
 & $O{+}I{+}D$ & 0.16 & 0.12 & 0.25 & 0.06 & 0.06 & 0.02 & 1.00 & 0.00 \\
\midrule
\multirow{8}{*}{\begin{tabular}[c]{@{}c@{}}gemini-3.5\\-flash\end{tabular}}
 & Baseline & 0.08 & 0.14 & 0.11 & 0.11 & 0.13 & 0.17 & 1.00 & 0.00 \\
 & $O$ & 0.13 & 0.19 & 0.40 & 0.32 & 0.14 & 0.17 & 0.30 & 0.33 \\
 & $I$ & 0.02 & 0.10 & 0.26 & 0.08 & 0.09 & 0.10 & 0.50 & 0.00 \\
 & $D$ & 0.13 & 0.18 & 0.35 & 0.32 & 0.11 & 0.14 & 0.15 & 0.30 \\
 & $O{+}I$ & 0.07 & 0.14 & 0.21 & 0.13 & 0.08 & 0.08 & 0.62 & 0.00 \\
 & $O{+}D$ & 0.14 & 0.20 & 0.39 & 0.34 & 0.13 & 0.20 & 0.50 & 0.05 \\
 & $I{+}D$ & 0.05 & 0.12 & 0.28 & 0.17 & 0.10 & 0.10 & 0.60 & 0.13 \\
 & $O{+}I{+}D$ & 0.13 & 0.17 & 0.43 & 0.23 & 0.12 & 0.07 & 0.59 & 0.09 \\
\midrule
\multirow{8}{*}{\begin{tabular}[c]{@{}c@{}}gpt-5.5\end{tabular}}
 & Baseline & 0.00 & 0.05 & 0.05 & 0.05 & 0.03 & 0.32 & 1.00 & 0.00 \\
 & $O$ & 0.03 & 0.09 & 0.21 & 0.14 & 0.04 & 0.10 & 1.00 & 0.00 \\
 & $I$ & 0.00 & 0.05 & 0.15 & 0.00 & 0.02 & 0.24 & 1.00 & 0.00 \\
 & $D$ & 0.01 & 0.10 & 0.17 & 0.10 & 0.02 & 0.17 & 1.00 & 0.00 \\
 & $O{+}I$ & 0.00 & 0.07 & 0.17 & 0.02 & 0.01 & 0.08 & 0.75 & 0.00 \\
 & $O{+}D$ & 0.08 & 0.12 & 0.33 & 0.20 & 0.07 & 0.19 & 0.43 & 0.00 \\
 & $I{+}D$ & 0.00 & 0.06 & 0.15 & 0.15 & 0.01 & 0.17 & 0.75 & 0.00 \\
 & $O{+}I{+}D$ & 0.04 & 0.07 & 0.17 & 0.04 & 0.01 & 0.08 & 1.00 & 0.00 \\
\midrule
\multicolumn{10}{@{}p{0.96\textwidth}@{}}{
\footnotesize
\textit{Abbreviations:}
$O$ = information omission;
$I$ = contradiction injection;
$D$ = evidence delay;
NFR = Numerical Fabrication Ratio;
UFR = Unsubstantiated Fact Ratio;
HPR = Hallucination Persistence Ratio;
HCCR = Hallucination Cross-Contamination Ratio;
DHDR = Definitive Hallucination Dependency Ratio;
CHO = Critical Hallucination Omission;
CIHSR = Contradiction-Induced Hallucination Suppression Ratio;
CIHGR = Contradiction-Induced Hallucination Generation Ratio.
} \\
\end{tabular}
\end{adjustbox}
\endgroup
\end{table}

\begin{table}[t]
\centering
\begingroup
\renewcommand{\arraystretch}{0.78}
\caption{Hallucination propagation evaluation results on the \textbf{agent track} under different information-flow stress conditions. Metrics are grouped by the four hallucination dimensions.}
\label{tab:agent_hallucination_results}
\begin{adjustbox}{max width=\textwidth,max totalheight=0.72\textheight}
\begin{tabular}{@{}cc*{8}{c}@{}}
\toprule
\multirow{2}{*}{\textbf{Model}}
& \multirow{2}{*}{\textbf{Stress}}
& \multicolumn{2}{c}{\textbf{Initiation}}
& \multicolumn{2}{c}{\textbf{Propagation}}
& \multicolumn{2}{c}{\textbf{Anchoring}}
& \multicolumn{2}{c}{\textbf{Contra. Inter.}}
\\
\cmidrule(lr){3-4}
\cmidrule(lr){5-6}
\cmidrule(lr){7-8}
\cmidrule(lr){9-10}
&
& \textbf{NFR}$\downarrow$
& \textbf{UFR}$\downarrow$
& \textbf{HPR}$\downarrow$
& \textbf{HCCR}$\downarrow$
& \textbf{DHDR}$\downarrow$
& \textbf{CHO}$\downarrow$
& \textbf{CIHSR}$\uparrow$
& \textbf{CIHGR}$\downarrow$ \\
\midrule
\multirow{8}{*}{\begin{tabular}[c]{@{}c@{}}kimi-k2.6\end{tabular}}
 & Baseline & 0.16 & 0.17 & 0.24 & 0.21 & 0.09 & 0.12 & 1.00 & 0.00 \\
 & $O$ & 0.12 & 0.18 & 0.41 & 0.24 & 0.11 & 0.19 & 0.67 & 0.00 \\
 & $I$ & 0.11 & 0.14 & 0.24 & 0.07 & 0.10 & 0.07 & 1.00 & 0.06 \\
 & $D$ & 0.12 & 0.14 & 0.28 & 0.13 & 0.08 & 0.10 & 0.50 & 0.00 \\
 & $O{+}I$ & 0.10 & 0.13 & 0.31 & 0.11 & 0.07 & 0.11 & 0.67 & 0.02 \\
 & $O{+}D$ & 0.16 & 0.17 & 0.37 & 0.16 & 0.10 & 0.06 & 0.75 & 0.07 \\
 & $I{+}D$ & 0.13 & 0.13 & 0.22 & 0.13 & 0.06 & 0.17 & 1.00 & 0.04 \\
 & $O{+}I{+}D$ & 0.06 & 0.11 & 0.20 & 0.03 & 0.03 & 0.05 & 1.00 & 0.00 \\
\midrule
\multirow{8}{*}{\begin{tabular}[c]{@{}c@{}}claude\\opus-4-7\end{tabular}}
 & Baseline & 0.14 & 0.16 & 0.23 & 0.10 & 0.08 & 0.02 & 1.00 & 0.00 \\
 & $O$ & 0.17 & 0.17 & 0.35 & 0.20 & 0.10 & 0.11 & 1.00 & 0.00 \\
 & $I$ & 0.13 & 0.12 & 0.17 & 0.03 & 0.06 & 0.04 & 1.00 & 0.00 \\
 & $D$ & 0.14 & 0.16 & 0.39 & 0.16 & 0.10 & 0.07 & 0.00 & 0.00 \\
 & $O{+}I$ & 0.15 & 0.14 & 0.25 & 0.09 & 0.06 & 0.09 & 0.75 & 0.10 \\
 & $O{+}D$ & 0.15 & 0.16 & 0.39 & 0.17 & 0.10 & 0.03 & 0.33 & 0.00 \\
 & $I{+}D$ & 0.12 & 0.13 & 0.22 & 0.07 & 0.05 & 0.09 & 1.00 & 0.04 \\
 & $O{+}I{+}D$ & 0.14 & 0.12 & 0.25 & 0.04 & 0.04 & 0.02 & 0.88 & 0.00 \\
\midrule
\multirow{8}{*}{\begin{tabular}[c]{@{}c@{}}grok-4.20\\beta-0309-reasoning\end{tabular}}
 & Baseline & 0.17 & 0.19 & 0.25 & 0.27 & 0.14 & 0.20 & 1.00 & 0.00 \\
 & $O$ & 0.14 & 0.20 & 0.40 & 0.28 & 0.11 & 0.26 & 1.00 & 0.00 \\
 & $I$ & 0.15 & 0.14 & 0.26 & 0.12 & 0.09 & 0.09 & 0.67 & 0.06 \\
 & $D$ & 0.16 & 0.18 & 0.34 & 0.28 & 0.10 & 0.15 & 0.50 & 0.00 \\
 & $O{+}I$ & 0.16 & 0.15 & 0.30 & 0.14 & 0.08 & 0.14 & 1.00 & 0.00 \\
 & $O{+}D$ & 0.16 & 0.21 & 0.37 & 0.19 & 0.10 & 0.17 & 0.67 & 0.00 \\
 & $I{+}D$ & 0.14 & 0.13 & 0.25 & 0.16 & 0.05 & 0.18 & 1.00 & 0.04 \\
 & $O{+}I{+}D$ & 0.14 & 0.14 & 0.47 & 0.14 & 0.06 & 0.16 & 0.73 & 0.00 \\
\midrule
\multicolumn{10}{@{}p{0.96\textwidth}@{}}{
\footnotesize
\textit{Abbreviations:}
$O$ = information omission;
$I$ = contradiction injection;
$D$ = evidence delay;
NFR = Numerical Fabrication Ratio;
UFR = Unsubstantiated Fact Ratio;
HPR = Hallucination Persistence Ratio;
HCCR = Hallucination Cross-Contamination Ratio;
DHDR = Definitive Hallucination Dependency Ratio;
CHO = Critical Hallucination Omission;
CIHSR = Contradiction-Induced Hallucination Suppression Ratio;
CIHGR = Contradiction-Induced Hallucination Generation Ratio.
} \\
\end{tabular}
\end{adjustbox}
\endgroup
\end{table}

\end{document}